\documentclass[a4paper,11pt]{article}

\usepackage{amsmath,amssymb}
\usepackage{bm}
\usepackage{upgreek}
\usepackage{textcomp}
\usepackage{url}
\usepackage{rotating}
\usepackage{gensymb}
\usepackage{natbib}
\usepackage[english]{babel}
\usepackage[T1]{fontenc}
\usepackage{float}
\usepackage[top=1in, bottom=1.25in, left=1.25in, right=1.25in]{geometry}

\def\acknowledgments{\vskip12pt\noindent{\bf
Acknowledgments\vrule depth 6pt
width0pt\relax}\\*\noindent\ignorespaces}

\title{Training-image based geostatistical inversion using a spatial generative adversarial neural network}

\author{Eric Laloy\thanks{Belgian Nuclear Research Centre, Engineered and Geosystems Analysis. Email: {\tt elaloy@sckcen.be}}, Romain H\'erault\thanks{Normandie Univ, UNIROUEN, UNIHAVRE, INSA Rouen, LITIS, 76000 Rouen, France}, Diederik Jacques\thanks{Belgian Nuclear Research Centre, Engineered and Geosystems Analysis}, and Niklas Linde\thanks{Applied and Environmental Geophysics Group, Institute of Earth Sciences, University of Lausanne}}
\date{December 31, 2017}
\begin{document}

\maketitle

\begin{abstract}
Probabilistic inversion within a multiple-point statistics framework is often computationally prohibitive for high-dimensional problems. To partly address this, we introduce and evaluate a new training-image based inversion approach for complex geologic media. Our approach relies on a deep neural network of the generative adversarial network (GAN) type. After training using a training image (TI), our proposed spatial GAN (SGAN) can quickly generate 2D and 3D unconditional realizations. A key characteristic of our SGAN is that it defines a (very) low-dimensional parameterization, thereby allowing for efficient probabilistic inversion using state-of-the-art Markov chain Monte Carlo (MCMC) methods. In addition, available direct conditioning data can be incorporated within the inversion. Several 2D and 3D categorical TIs are first used to analyze the performance of our SGAN for unconditional geostatistical simulation. Training our deep network can take several hours. After training, realizations containing a few millions of pixels/voxels can be produced in a matter of seconds. This makes it especially useful for simulating many thousands of realizations (e.g., for MCMC inversion) as the relative cost of the training per realization diminishes with the considered number of realizations. Synthetic inversion case studies involving 2D steady-state flow and 3D transient hydraulic tomography with and without direct conditioning data are used to illustrate the effectiveness of our proposed SGAN-based inversion. For the 2D case, the inversion rapidly explores the posterior model distribution. For the 3D case, the inversion recovers model realizations that fit the data close to the target level and visually resemble the true model well.
\end{abstract}

\section{Introduction}
\label{intro}

Capturing subsurface complexity and connectivity is of utmost importance for reliable groundwater modeling and uncertainty quantification \citep[e.g.,][]{Mariethoz-Caers2014,Linde2015}. Subsurface flow and transport modeling studies often rely on inversion methodologies to derive subsurface structures that are consistent with both available prior information and indirect measurements of one or more state variables such as hydraulic head or concentration. When it is unrealistic to assume that the subsurface is multi-Gaussian \citep[e.g.,][]{GomezHernandez-Wen1998, Journel-Zhang2006}, one popular solution is to resort to multiple-point statistics (MPS) simulation \citep[e.g.,][]{Strebelle2002,Mariethoz2010,Li2016,Tahmasebi-Sahimi2016a,Tahmasebi-Sahimi2016b, Tahmasebi2017}. MPS techniques generate model realizations that honor a prior model which is determined by a training image (TI). The TI is a large gridded 2D or 3D unconditional representation of the expected target spatial field that can be either continuous or categorical (e.g., geologic facies image). Despite recent advances \citep[e.g.,][]{Zahner2016,Laloy2016,Jaeglli2017}, MPS-based probabilistic inversion is slow  and the exploration of the posterior density is often incomplete. Different model reduction techniques \citep[e.g.,][and references therein]{Vo-Durlofsky2014} exist but they generally do not provide model realizations that agree well with the MPS characteristics of the TIs used (i.e., geological realism is lost). The problem of how to effectively address inversion that honor the information content of complex geological priors is thus a largely open and important research question.

Deep learning is currently transforming science and associated industries \citep[][]{Goodfellow2016}. Deep learning can be thought of as a class of mathematical models that automatically learn a new parametric representation of the data that are given to them and, for some specific models, enable inverse mapping from the learned representation to the original data space. Note that the term ``data" is here employed in a wide sense, and can thus include images, text, etc. More specifically, a TI is considered data in this terminology. Technically speaking, ``deep" refers to the number of layers in a so-called (artificial) neural network. Yet it also more broadly relates to increased model flexibility and complexity. A deep model takes raw data and constructs complex representations that are expressed in terms of other, simpler representations \citep[e.g.,][]{Goodfellow2016}. 

Recently, \citet{Laloy2017} have proposed a deep neural network model of the variational autoencoder (VAE) type that is able to build a low-dimensional parameterization of complex binary geologic prior models. They then performed probabilistic inversion within the resulting low-dimensional parameter space. This approach was shown to outperform state-of-the-art MPS-based inversion by sequential geostatistical resimulation \citep[SGR, also known as iterative spatial resampling and sequential Gibbs sampling, see, e.g.,][]{Mariethoz2010b, Hansen2012,Laloy2016, Zahner2016} for the considered problem. Nevertheless, the VAE-based inversion approach by \citet{Laloy2017} suffers from three limitations: (1) it needs a large set (tens of thousands) of training images, (2) it is applicable to binary geologic models only and (3) it defines a global model compression in which each of the low-dimensional (latent) parameters influences the whole spatial model domain  (i.e., the original model space that might correspond to a gridded 2D or 3D field). In this work, we address these issues by using a totally different type of neural network called spatial generative adversarial neural network (SGAN).

Generative neural networks are deep networks that can be trained to stochastically generate data with similar properties as a training dataset \citep[e.g.,][]{Goodfellow2016}. In the following simulation and inversion examples, we will for clarity use the term ``model" instead of ``data" when referring to a representation of subsurface property fields. Lately, generative adversarial networks \citep[GAN,][]{Goodfellow2014} have become increasingly popular in the field of deep learning. The main specificity of GANs is that they are learned by creating a competition between the actual generative model or ``generator" and a discriminator in a zero-sum game framework \citep{Goodfellow2014}, in which these components are learned jointly. The discriminator tries to distinguish between real (training) data and generated data (i.e., to detect imperfections in the realizations), and the generator aims at fooling the discriminator, thereby, improving the realizations. A large variety of GANs have recently been proposed to generate different types of high-dimensional data, mostly images \citep[e.g.,][]{Radford2016,Jetchev2016}, but also sound \citep[e.g.,][]{Yang2017} and text \citep[e.g.,][]{Liang2017} to list a few application areas. 

In this work, we introduce the idea and demonstrate the performance of using a spatial GAN (SGAN) for MPS-based geostatistical inversion in 2D and 3D spatial domains. We propose a 3D extension of the original 2D SGAN by \citet{Jetchev2016}. To enable efficient Markov chain Monte Carlo (MCMC) inversion, we take advantage of the fact that, similarly to the deep variational autoencoder (VAE) used by \citet{Laloy2017}, the standard GAN architecture defines a low-dimensional ``latent" representation of the original high-dimensional data (i.e., spatial subsurface property fields in our case). In such an architecture, the high-dimensional data realizations are obtained by propagating a low-dimensional random noise vector through the network. The inversion parameters can thus be defined by and state-of-the-art MCMC sampling can be performed within this low-dimensional latent space. Compared to our previous work with a VAE \citep{Laloy2017}, our SGAN has the following advantages: (1) training the network is performed using a single TI only, (2) it is fully 3D, (3) it can handle both binary and multi-categorical data (geological facies in our case), (4) it is designed such that each dimension of the latent space influences a specific part of the high-dimensional domain, and (5) it results into an even more compact representation of high-dimensional 3D domains with a dimensionality reduction that can exceed 4 orders of magnitude.

The latent (model) representation learned by the SGAN defines a low-dimensional uniform parameter space. Random draws from this low-dimensional space can be fed to the SGAN generator to create spatial model realizations with similar spatial statistics as those found in the TI. From an inversion perspective, it is very attractive to reduce an inverse problem with tens (or hundreds) of thousands of unknowns with complex spatial priors to the inference of an independent low-dimensional latent space. For a given hydrogeological forward model, the inversion performance will be largely dependent on (1) the time needed for training the network and performing the mapping from the latent space to the spatial domain needed for forward model simulations (ie., the time required by the generator to produce a realization) and (2) the ability of the generator to produce geostatistical realizations with similar MPS characteristics as the TI. Indeed, high-quality inversion results can only be achieved if the realizations are of sufficiently high quality in the first place. This implies that it is necessary to investigate the ability of our SGAN to produce unconditional geostatistical realizations. As of direct conditioning data, we demonstrate that they can be incorporated within the inversion.

The geostatistical simulation component of our paper has similarities with the independently developed study by \citet{Mosser2017}, in which a slightly different type of GAN called deep convolutional generative adversarial network \citep[DCGAN,][]{Radford2016} is used for 3D generation of binary porous systems. The main differences between our work and that of Mosser and coworkers are as follows. First, for our SGAN each dimension of the low-dimensional latent representation influences only a given region of the original model domain, which is beneficial for inversion. Second, we consider more complex training images than the binary images investigated by \citet{Mosser2017}. Indeed, our used TIs show a much larger degree of connectivity and can be multi-categorical. Finally and most importantly, we use our proposed SGAN to perform geostatistical inversion.

This paper is organized as follows. Section \ref{methods} presents the different elements needed for (unconditional) geostatistical simulation and inversion using a SGAN architecture. This is followed in section \ref{results_sim} with a performance analysis of the geostatistical simulation capabilities using several 2D and 3D categorical TIs. As discussed above, it is necessary to assess the quality of the produced geostatistical realizations prior to probabilisitc geostatistical inversion. Synthetic 2D and 3D experiments involving both steady-state and transient groundwater flow are then used in section \ref{results_inv} to demonstrate our proposed inverse method's capabilities. In section \ref{discussion}, we discuss the advantages and limitations of our method and outline possible future developments. Finally, section \ref{conclusion} concludes with a summary of the most important findings.

\section{Methods}
\label{methods}

\subsection{Spatial Generative Adversarial Network Architecture}

Our proposed nerual network is a 3D extension of the 2D SGAN by \citet{Jetchev2016}. This type of neural network is a deep neural network (DNN) that is built of direct and transposed convolutional layers only. Basically, neural networks approximate the (potentially complex) relationships between input, $\textbf{x}$, and output, $\textbf{y}$, data vectors by combining many computational units called ``neurons". In addition, the classical DNN architecture stacks successive layers of neurons. A typical formulation of the neuron is given by
\begin{equation}
	h\left(\textbf{x}\right) =f\left(\langle \textbf{x}, \textbf{w} \rangle+ b \right),
	\label{sgan0}
\end{equation}
where $h\left(\cdot \right)$ signifies the scalar output of the neuron, $f\left(\cdot \right)$ is a nonlinear function that is called the ``activation function", $\langle\cdot,\cdot\rangle$ denotes the scalar product, $\textbf{w} = \left[w_1, \cdots, w_N\right]$ is a set of weights of the same dimension, $N$, as $\textbf{x}$ and $b$ represents the bias associated with the neuron. To be useful, a DNN must be trained or ``learned". In the learning process the values in $\textbf{w}$ and $b$ are optimized for each neuron such that the DNN performs a prespecified task as well as possible. When $f\left(\cdot \right)$ is differentiable, $\textbf{w}$ and $b$ can be optimized by gradient descent. Typical forms of $f\left(\cdot \right)$ include the rectified linear unit (ReLU), sigmoid function and hyperbolic tangent function \citep[e.g.,][]{Goodfellow2016}.

The convolutional layer forms the fundamental building block of the convolutional neural network (CNN) type of architecture. The convolutional operator has gained widespread use for image processing because it explicitly considers the spatial structure in the input data. When the input is a 2D image (with possibly 3 channels for a RGB-color image), a convolutional layer, $\textbf{h}$, is constructed by means of a series of $k = 1, \cdots, N_k$ small $N_i \times N_j$ filters, $\textbf{w}^k$, that convolve an input pixel, $X_{u,v}$ to $h^{k}_{u,v}$ as
\begin{equation}
	h^{k}_{u,v}\left(X_{u,v}\right) = f\left(\sum_{i = 1}^{N_i}\sum_{j = 1}^{N_j} w^{k}_{i,j}X_{u+i,v+j} + b_k \right),
	\label{sgan1}
\end{equation}
where a common choice for $f\left(\cdot \right)$ is the rectified linear unit (ReLU): $f\left(x \right) = \max\left(0,x\right)$. The resulting series of $N_k$ $\textbf{h}^k$ stacked ``feature maps" forms the convolutional layer $\textbf{h}$. Note that the neuron defined by equation (\ref{sgan1}) is connected to only a local region of the input image (or volume for the 3D formulation). This enforces encoding of spatially-local information at the level of one convolutional layer. Therefore, the larger the number of stacked feature maps, $N_k$, the richer the representation of the input data. There are other important convolution parameters that, for brevity, we do not discuss herein. These are the ``stride" parameter which controls the overlapping between successive moves in the forward pass of a given filter, the `padding" parameter that defines padding of the borders of the input images or volumes for size preservation, and the ``pooling" parameter which involves non-linear down-sampling to reduce the size of a convolutional layer and thereby prevents overfitting. For further information on CNN layers, we refer the reader to \citet{Goodfellow2016} and online tutorials\footnote{For instance: \url{http://deeplearning.net/tutorial/} and \url{http://cs231n.github.io/convolutional-networks/}.}.

\begin{figure}[h!]
\noindent\hspace{0cm}\includegraphics[width=35pc]{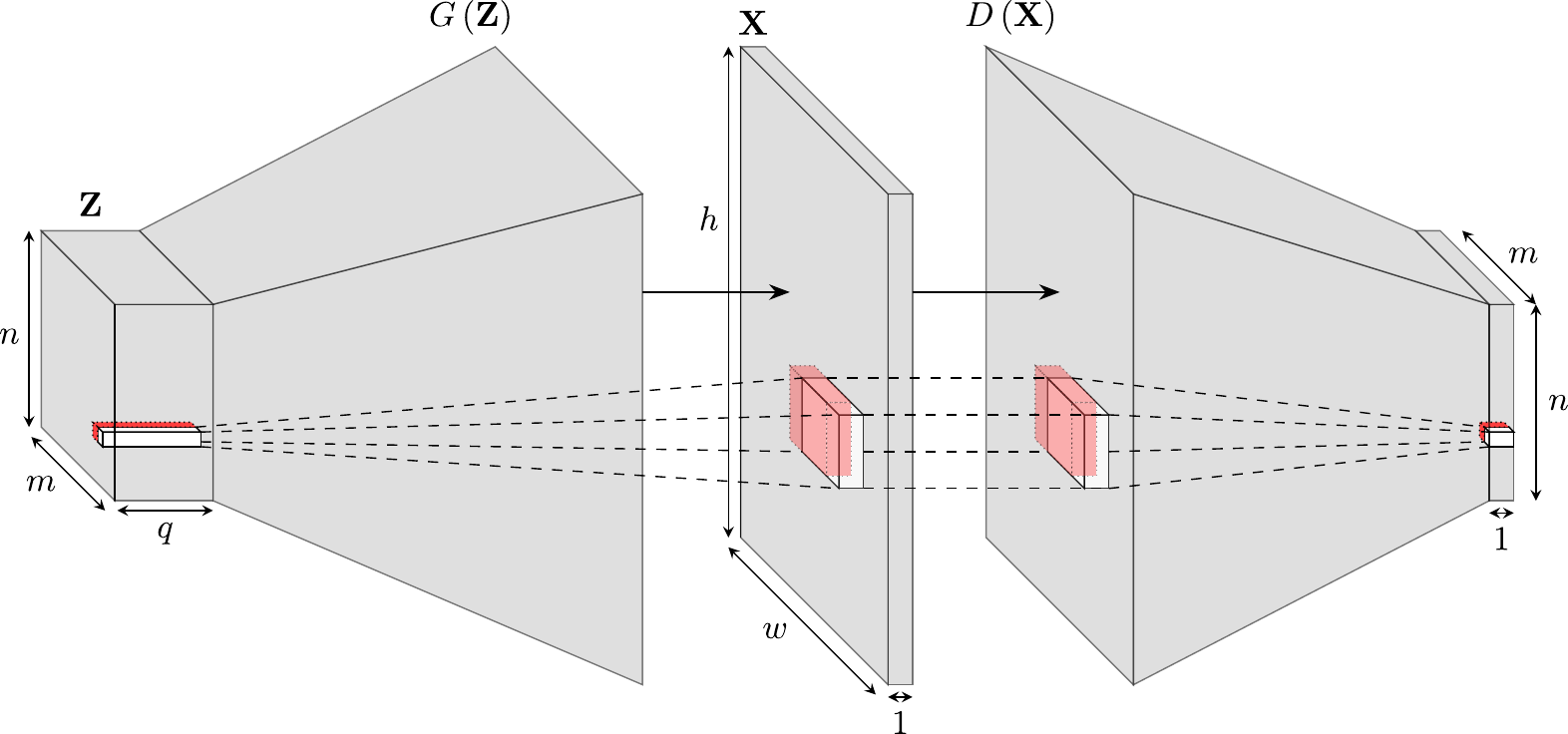}
\caption{Illustration of the SGAN structure for the 2D case. A input $m \times n \times q$ $\textbf{Z}$ array of uniform variables $\sim U\left(-1,1\right)$ is propagated through the generator, $G\left(\textbf{Z}\right)$, to produce a grayscale image, $\textbf{X}$. During learning, the generated $\textbf{X}$ and true $\textbf{X}$ iteratively enter the discriminator, $D\left(\textbf{X}\right)$. The latter produces a $m \times n$ array of probabilities, the mean of which is taken as the probability that the incoming $\textbf{X}$ belongs to the TI. Generator and discriminator are made of stacked transposed and non-transposed convolutional layers, respectively. The network structure for the 3D case obeys the same logic but it cannot easily be represented.}
\label{fig1}
\end{figure}

Figure \ref{fig1} depicts the main SGAN architecture for the generation of  2D grayscale images. The 3D case obeys the same principles but it cannot easily be represented visually. The low-dimensional input or latent space, $\textbf{Z}$, has a spatial structure and follows a bounded uniform distribution, $\textbf{Z} \sim U\left(-1,1\right)$. In the 2D case, $\textbf{Z}$ is a 3D array of size $m \times n \times q$. Each $Z_{\left(i,j,\cdot\right)}$ with $i = 1, \cdots,m; j = 1, \cdots,n$ controls a specific region of the generated full scale $w \times h$ image, $\textbf{X}$. In addition, these specific regions partially overlap. This is illustrated for the $1 \times 1 \times q$ white-colored component of $\textbf{Z}$ in Figure \ref{fig1}. The third dimension, $q$, is not related to a spatial location but allows for additional flexibility in the data representation encoded by each $q$-dimensional element $Z_{\left(i,j,k = 1 \cdots q\right)}$. For 3D grayscale image generation, $\textbf{Z} \sim U\left(-1,1\right)$ becomes a 4D array of size $m \times n \times o \times q$  where the dimensions $m$, $n$ and $o$ now correspond to specific regions of the generated full scale volume of size $w \times h \times l$. At generation time, the sampled  $\textbf{Z}$ array enters the generator, $G\left(\textbf{Z}\right)$ to produce a (2D or 3D) grayscale image, $\textbf{X}$ (Figure \ref{fig1}). $G\left(\textbf{Z}\right)$ is made from a stack of transposed 2D or 3D convolutional layers \citep[see][for details on the transpose operation for convolutional layers]{Dumoulin-Visin2016}. At discrimination time, that is, during training, either the generated image or a true image formed by a fraction of the TI (interchangeably called $\textbf{X}$ for now) is processed through the discriminator, $D\left(\textbf{X}\right)$, to output a $m \times n$ (2D case) or $m \times n \times o$ (3D case) field of probabilities for fake/real images (see also section \ref{training}). 

The computational cost incurred by training a deep generative model largely depends on the size of the model realization domain. For stationary fields, the pure convolutional nature of the original 2D SGAN and our 3D extension makes it possible to train the network at reduced computational cost using a relatively small realization domain and then use the trained network to generate larger realizations that match the spatial statistics found in the training data. The actual working values behind the notions of small and large realization domains mainly depend on the relative size of the relevant patterns in the stationary TI, together with hardware capabilities. For instance, for the binary channelized TI depicted in Figure \ref{fig2}a, training using a $353 \times 353$ realization domain (see section \ref{2d_gen}) allows for generating realizations of size $3553 \times 3553$ that are of same quality as the $609 \times 609$ realizations analyzed in section \ref{2d_gen} (not shown). For a square ($w \times h$) or cubic ($w \times h \times l$) generation domain, the relationship  between $z_{\rm x}= m = n = o$ and $x_{\rm x} = w = h = l$ is given by
\begin{equation}
	x_{\rm x} = \left[z_{\rm x} - 1\right]2^{dp} + 1,
	\label{sgan2}
\end{equation}
where $dp$ is the number of stacked convolutional layers in $G\left(\textbf{Z}\right)$. This allows for a strong dimensionality reduction of the TI space. We refer to \citet{Jetchev2016} for more details on the (2D) SGAN architecture. 

For all our experiments, we used squared or cubic filters with side length ($N_i$ in equation (\ref{sgan1})) in the range 3-9 depending on the case study and the same activation functions ($f\left(\cdot \right)$ in equation (\ref{sgan1})) as in \citet{Jetchev2016}. These are: ReLU (all but last layer) and hypertangent (last layer) for $G\left(\textbf{Z}\right)$ and so-called ``leaky" ReLU of the form $f\left(x \right) = \max\left(0.2x,x\right)$ (all but last layer) and sigmoid (last layer) for $D\left(\textbf{X}\right)$. Moreover, we always considered a stack of $dp = 5$ convolutional layers for both $G\left(\textbf{Z}\right)$ and $D\left(\textbf{X}\right)$ with the following number of feature maps ($N_k$ in equation (\ref{sgan1})) from $\textbf{Z}$ to $\textbf{X}$: $\left[512, 256, 128, 64, 1\right]$. 

Using  $dp = 5$ together with $q=1$, as we do for 2 out of our 3 test cases, equation (\ref{sgan2}) predicts that 3D realizations of size $129 \times 129 \times 129$ can be generated by drawing as few as $5 \times 5 \times 5 = 125$ $Z$ variables from the $U\left(-1,1\right)$ distribution. This corresponds to a more than 4 orders of magnitude dimensionality reduction. Note that in this study, we never used a $q$ value larger than 3. For the above example, $q = 3$ would still lead to a more than 3 orders of magnitude dimensionality reduction.

Similarly as the original 2D SGAN, our 3D extension was implemented within the open-source LASAGNE Python software \citep{lasagne} which works on top of the open-source THEANO Python library \citep{theano}.

\subsection{Training The Generative Adversarial Network}
\label{training}
The specificity of the GAN architecture is that the generator and discriminator are trained (or ``learned") simultaneously with opposed goals. The discriminator, $D\left(\textbf{X}\right)$, is fed with samples from the ``real" training set, which from now on will be referred to as $\textbf{X}$ with distribution $p_{\rm data}\left(\textbf{X}\right)$, and ``fake" samples (i.e., realizations) created by the generator: $\hat{\textbf{X}} = G\left(\textbf{Z}\right)$. The real samples $\textbf{X}$ are a set of patches randomly cut from the TI. The discriminator tries to distinguish between $\textbf{X}$ and $\hat{\textbf{X}}$ by computing for each received sample the probability that it belongs to $p_{\rm data}\left(\textbf{X}\right)$. In contrast, the generator, $G\left(\textbf{Z}\right)$ aims at fooling $D\left(\cdot\right)$ into labeling $\hat{\textbf{X}}$ as a sample from $p_{\rm data}\left(\textbf{X}\right)$ and thus achieving a mean $D\left(\hat{\textbf{X}}\right) = D\left(G\left(\textbf{Z}\right)\right)$ close to 1 \citep{Goodfellow2014}. Mathematically, this translates into the following minimization-maximization loss function
\begin{equation}
	\min_{G\left(\cdot\right)} \max_{D\left(\cdot\right)}\left\{\mathbb{E}_{\textbf{X}\sim p_{\rm data}\left(\textbf{X}\right)}\left[\log\left(D\left(\textbf{X}\right)\right)\right] +\mathbb{E}_{\textbf{Z}\sim p_{\textbf{Z}}\left(\textbf{Z}\right)}\left[\log\left(1-D\left(G\left(\textbf{Z}\right)\right)\right)\right]\right\}.
	\label{sgan3}
\end{equation}

In practice, the minimization-maximization problem in equation (\ref{sgan3}) is solved in two consecutive steps. First one minimizes
\begin{equation}
	\mathcal{L}^{D} = -\mathbb{E}_{\textbf{X}\sim p_{\rm data}\left(\textbf{X}\right)}\left[\log\left(D\left(\textbf{X}\right)\right)\right] - \mathbb{E}_{\textbf{Z}\sim p_{\textbf{Z}}\left(\textbf{Z}\right)}\left[\log\left(1-D\left(G\left(\textbf{Z}\right)\right)\right)\right]
	\label{sgan4}
\end{equation}
while keeping the parameters of $G\left(\textbf{Z}\right)$ (that is, the $\textbf{W}$ and $\textbf{b}$ of each layer of  $G\left(\textbf{Z}\right)$) fixed to current values. Second, one minimizes
\begin{equation}
	\mathcal{L}^{G} = \mathbb{E}_{\textbf{Z}\sim p_{\textbf{Z}}\left(\textbf{Z}\right)}\left[\log\left(1-D\left(G\left(\textbf{Z}\right)\right)\right)\right]
	\label{sgan5}
\end{equation}
with the parameters of $D\left(\cdot\right)$ fixed to current values.

For numerical stability, $l_2$-norm regularization operators for both the generator and discriminator weights are added to equations (\ref{sgan4}) and (\ref{sgan5}). Furthermore, early in the training process when $G\left(\cdot\right)$ is poor and $D\left(\cdot\right)$ very efficient the quantity $\left[\log\left(1-G\left(\textbf{Z}\right)\right)\right]$ in equation (\ref{sgan5}) is known to be prone to saturation. That is, $\left[\log\left(1-G\left(\textbf{Z}\right)\right)\right]$ becomes close to zero and thus does not provide enough gradient for $G\left(\cdot\right)$ to be trained. To avoid this problem, \citet{Goodfellow2014} proposed to replace $\left[\log\left(1-G\left(\textbf{Z}\right)\right)\right]$ by $-\left[\log\left(G\left(\textbf{Z}\right)\right)\right]$ in equation (\ref{sgan5}). Approximating the expectations by sums, this leads (for the 3D case) to the following formulations 
	\begin{equation}
	\mathcal{L}^{D} = -\frac{1}{mno}\sum_{i=1}^{m}\sum_{j=1}^{n}\sum_{k=1}^{o}\log\left(D_{ijk}\left(\textbf{X}\right)\right) -\frac{1}{mno}\sum_{i=1}^{m}\sum_{j=1}^{n}\sum_{k=1}^{o}\log\left(1-D_{ijk}\left(G\left(\textbf{Z}\right)\right)\right) +\alpha\left\|\bm{\Upomega}_{\rm D}\right\|^2_{2}
	\label{sgan6}
	\end{equation}
and
\begin{equation}
	\mathcal{L}^{G} = -\frac{1}{mno}\sum_{i=1}^{m}\sum_{j=1}^{n}\sum_{k=1}^{o}\log\left(D_{ijk}\left(G\left(\textbf{Z}\right)\right)\right) +\alpha\left\|\bm{\Upomega}_{\rm G}\right\|^2_{2}
	\label{sgan7}
\end{equation}

where $\bm{\Upomega}_{\rm D}$ and $\bm{\Upomega}_{\rm G}$ contain the discriminator and generator weights, respectively, and $\alpha$ is a weight parameter that we set to $1 \times 10^{-5}$.

Since both $G\left(\cdot\right)$ and $D\left(\cdot\right)$ are differentiable, equations (\ref{sgan6}-\ref{sgan7}) can be minimized by stochastic gradient descent (i.e., gradient descent using a series of mini-batches rather than all the data at once) together with back propagation. This means that the loss function derivative is propagated backwards throughout the network using the chain rule, in order to update the parameters. Various stochastic gradient descent algorithms are available. In this work, we used the adaptive moment estimation (ADAM) algorithm which has been proven efficient for different types of deep networks \citep{Kingma-Ba2015}.

The minimization of equations (\ref{sgan6}-\ref{sgan7}) was performed on a GPU Tesla K40 and training the SGAN for 50 epochs (full cycles of the stochastic gradient descent) took between 3 and 12 hours, depending mostly on the size of the model domain used for learning (from $129 \times 129$ to $97 \times 97 \times 97$) and the chosen number of patches in a training batch (from 25 to 64 depending on GPU memory availability).

Training of GANs is notoriously unstable, in the sense that image generation quality may not necessarily increase as training progresses \citep[e.g.,][]{Salimans2016,Sonderby2016,Mosser2017}. To mitigate this unstable behavior, we followed \citet{Sonderby2016} and added a Gaussian white noise with standard deviation of 0.1 to the input layer of $D\left(\cdot\right)$. Notwithstanding, this did not totally remove training instability (see supporting Figure S1). The final ``best" network was therefore chosen a posteriori for each case study by visual inspection of generation performance among the 50 trained networks associated with the achieved 50 training epochs. Whenever visual inspection was deemed insufficient, quantitative structural indicators were used as well (see section \ref{struct_indic}). This manual model selection step is arguably tedious yet inherent to the used GAN architecture. 

\subsection{Generation Quality Assessment}
\label{struct_indic}
Before proceeding to inversion, it is essential to verify that the unconditional realizations produced by randomly sampling the latent space of our SGAN are consistent with the TI. Indeed, we aim at producing inversion models that not only honor the data used, but also the spatial statistics of our selected TI. As mentionned above, the latter is only possible if the SGAN can generate high-quality geostatistical realizations. To complement visual inspection of randomly chosen realizations, a series of quantitative metrics were used to evaluate generation performance. For each case study, the following structural indicators were computed for (1) the TI and (2) a set 100 (2D case) or 25 (3D case) random realizations obtained by our SGAN:
\begin{enumerate}
	\item Two-point probability function \citep[PF,][]{Torquato-Stell1982},
	\item Two-point cluster function \citep[CF,][]{Torquato1988} which is also called connectivity function \citep{Pardo-Dowd2003},
	\item Fractions of the different facies.
\end{enumerate}

We refer the reader to the cited references for mathematical descriptions of these indicators. In short, the PF is the probability that 2 points separated by a given lag distance belong to the same facies while the CF is the probability that there exists a continuous path of the same facies between 2 points of the same facies separated by a given lag distance. Using the routines developed by \citet{Lemmens2017}, the PF, and CF were calculated for each facies along the $x$, $y$, and main diagonal, $d_{xy}$, directions for the 2D case. For the 3D case, the $z$ and main diagonal $d_{xz}$ and $d_{yz}$ directions were additionally considered, leading to a total of 6 metrics for each combination of indicator and facies.

\subsection{Bayesian inversion}
\label{methods_bayes}

A common representation of the forward problem is
\begin{equation}
\textbf{d} = F\left(\bm{\uptheta}\right) + \textbf{e},
\label{mcmc0}
\end{equation}
where $\textbf{d} = \left(d_1, \ldots, d_N \right) \in \mathbb{R}^N, N \geq 1$ is the measurement data, $F\left(\bm{\uptheta}\right)$ is a deterministic forward model with parameters $\bm{\uptheta}$ and the noise term $\textbf{e}$ lumps all sources of errors. 

In the Bayesian paradigm, parameters in $\bm{\uptheta}$ are viewed as random variables with a posterior pdf, $p\left(\bm{\uptheta} | \textbf{d} \right)$, given by
\begin{equation}
	p\left(\bm{\uptheta} | \textbf{d}  \right) = \frac{p \left(\bm{\uptheta}\right) p \left(\textbf{d} | \bm{\uptheta}\right)}{p \left( \textbf{d} \right)} \propto p\left(\bm{\uptheta}\right) L\left(\bm{\uptheta} | \textbf{d}\right),
	\label{mcmc1}
\end{equation}
where $L \left(\bm{\uptheta} | \textbf{d}\right) \equiv p \left(\textbf{d} | \bm{\uptheta}\right)$ signifies the likelihood function of $\bm{\uptheta}$. The normalization factor $p \left( \textbf{d} \right) = \int  p\left(\bm{\uptheta}\right) p\left(\textbf{d} | \bm{\uptheta}\right) d\bm{\uptheta}$ is not required for parameter inference when the parameter dimensionality is fixed. In the remainder of this paper, we will thus focus on the unnormalized density $p\left(\bm{\uptheta} | \textbf{d} \right) \propto p\left(\bm{\uptheta}\right) L\left(\bm{\uptheta} | \textbf{d}\right)$. 

To avoid numerical over- or underflow, it is convenient to work with the logarithm of $L \left(\bm{\uptheta} | \textbf{d}\right)$ (log-likelihood): $\ell\left(\bm{\uptheta} | \textbf{d}\right)$. If we assume $\textbf{e}$ to be normally distributed, uncorrelated and with known constant variance, $\sigma_e^2$, $\ell\left(\bm{\uptheta} | \textbf{d}\right)$ can be written as
\begin{equation}
	\ell\left(\bm{\uptheta} | \textbf{d}\right) = -\frac{N}{2}\log\left(2 \pi\right) - N\log\left(\sigma_e\right) -\frac{1}{2}\sigma_e^{-2}  \sum_{i = 1}^{N} \left[d_i - F_i\left(\bm{\uptheta}\right) \right]^2,
	\label{mcmc2}
\end{equation}
where the $F_i\left(\bm{\uptheta}\right)$ are the simulated responses that are compared the $i = 1, \cdots, N$ measurement data, $d_i$.

When $N_m$ direct conditioning point data, $\textbf{x}_m$, are available, we also try to enforce them by adding a second term to the log-likelihood
\begin{equation}
	\ell\left(\bm{\uptheta} | \textbf{x}_m\right) = -\frac{N_m}{2}\log\left(2 \pi\right) - N_m\log\left(\sigma_x\right) -\frac{1}{2}\sigma_x^{-2}  \sum_{i = 1}^{N_m} \left[x_{m,i} - G\left(\bm{\uptheta}\right)_i \right]^2,
	\label{mcmc3}
\end{equation}
and
\begin{equation}
	\ell\left(\bm{\uptheta} | \textbf{d}, \textbf{x}_m\right) = \ell\left(\bm{\uptheta} | \textbf{d}\right) + 	\ell\left(\bm{\uptheta} | \textbf{x}_m\right),
	\label{mcmc4}
\end{equation}
where $\sigma_x$ acts as a weigthing factor that balance the two components of $\ell\left(\bm{\uptheta} | \textbf{d}, \textbf{x}_m\right)$ and $G\left(\bm{\uptheta}\right)$ is the model realization created by feeding the SGAN generator, $G\left(\cdot\right)$, with $\bm{\uptheta}$. Moreover, notice that $\ell\left(\bm{\uptheta} | \textbf{x}_m\right)$ can also be viewed as a regularization prior for $\bm{\uptheta}$,  $\log\left[p\left(\bm{\uptheta} | \textbf{x}_m\right)\right]$.

As no analytical solution of $p\left(\bm{\uptheta} | \textbf{d}\right)$ or $p\left(\bm{\uptheta} | \textbf{d}, \textbf{x}_m\right)$ is available for the type of non-linear inverse problems considered herein, we sample from $p\left(\bm{\uptheta} | \textbf{d}\right)$ or $p\left(\bm{\uptheta} | \textbf{d}, \textbf{x}_m\right)$ by MCMC simulation \citep[see, e.g.,][]{Robert-Casella2004} with the DREAM$_{\rm \left(ZS\right)}$ algorithm \citep{Vrugt2009,Laloy-Vrugt2012}. Various studies in hydrology and geophysics (amongst others) have shown that DREAM$_{\rm \left(ZS\right)}$ can derive posterior distributions with 25-250 dimensions \citep[e.g.,][]{Laloy2012,Laloy2013,Linde-Vrugt2013,Laloy2015,Lochbuhler2015}. Since in this work the inversion is performed within the low-dimensional space defined by $\textbf{Z}$, we have $\bm{\uptheta} = \textbf{Z}$. Also, for our SGAN $p\left(\bm{\uptheta} | \textbf{d}\right)$ and $p\left(\bm{\uptheta} | \textbf{d}, \textbf{x}_m\right)$ reduce to $\ell\left(\bm{\uptheta} | \textbf{d}\right)$ and $\ell\left(\bm{\uptheta} | \textbf{d}, \textbf{x}_m\right)$, respectively, as the prior distribution of the latent parameters, $p\left(\bm{\uptheta}\right)$, is a uniform distribution. Given that we use symmetric model proposal distributions, only likelihood ratios are thus considered when calculating the acceptance probability in the MCMC.

Furthermore, we slightly modified the original DREAM$_{\rm \left(ZS\right)}$ algorithm by including a (vanishing) tempering of the likelihood function (i.e., starting with an inflated variance term in the likelihood function). Limited to the so-called burn-in \citep[see, e.g.,][]{Robert-Casella2004}, this tempering proved to be very useful for solving the considered inverse problems. 

\section{Geostatistical Simulation Results}
\label{results_sim}

As stated earlier, inversion is our primary objective. However, evaluating the quality of the geostatistial realizations produced by our approach is important as it will condition the ultimate quality of our inversion results.

\subsection{2D Models}
\label{2d_gen}

\begin{figure}[H]
\noindent\hspace{0cm}\includegraphics[width=35pc]{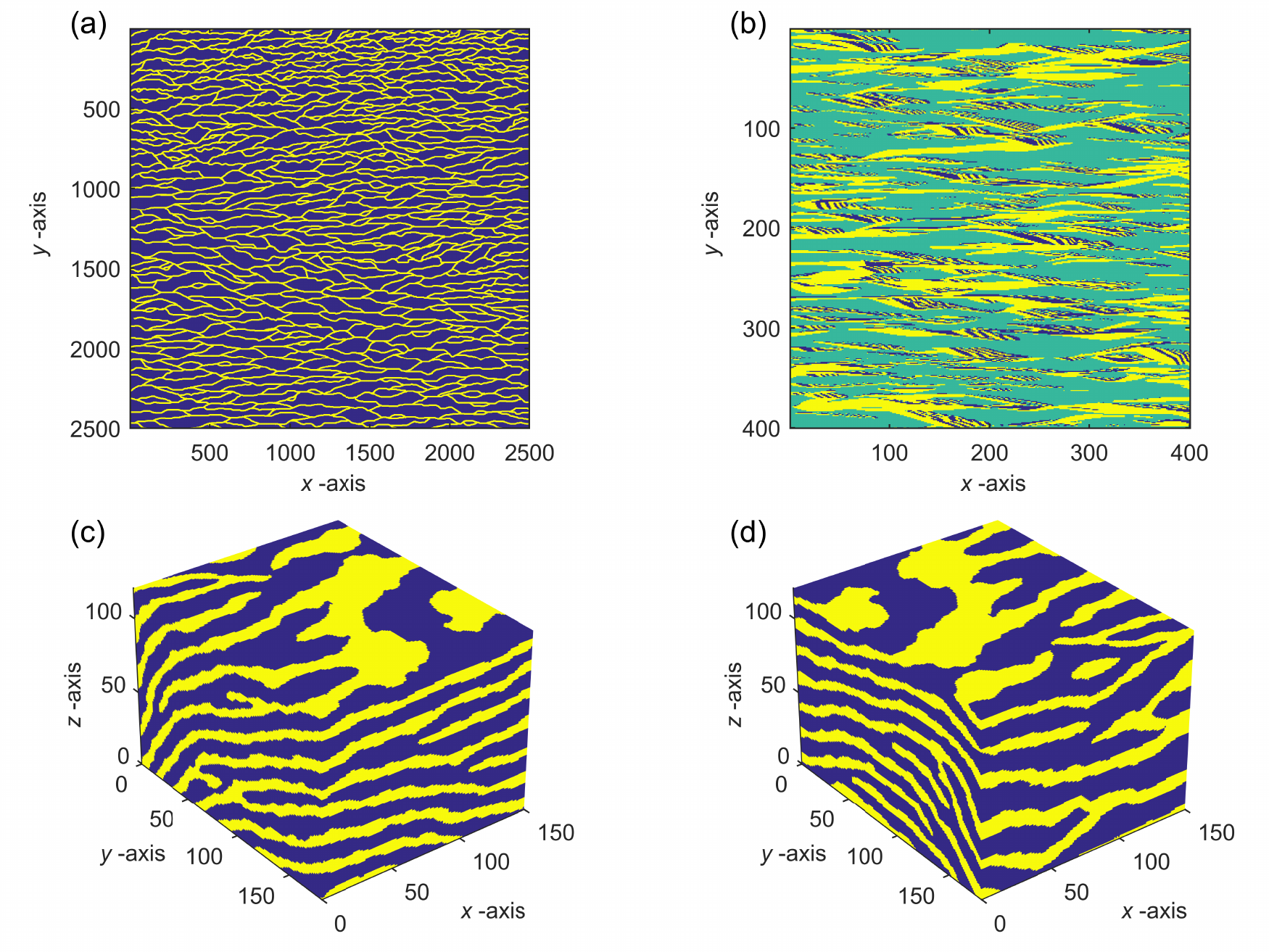}
\caption{Four training images (TIs) used for the considered geostatistical simulation tests. The TIs are (a) the $2500 \times 2500$ hand-made drawing inspired by Strebelle's TI by \citet{Zahner2016}, (b) the $400 \times 400$ braided river aquifer by \citet{Pirot2015}, (c) the $340 \times 200 \times 80$ Maules Creek aquifer (available at \protect\url{http://www.trainingimages.org/training-images-library.html} and (d) the $180 \times 150 \times 120$ categorical fold aquifer from \protect\url{http://www.trainingimages.org/training-images-library.html}. The 2D and 3D inverse case studies rely on the TIs depicted in (a) and (d), respectively.}
\label{fig2}
\end{figure}

Our first 2D test case considers the $2500 \times 2500$ binary channelized TI depicted in Figure \ref{fig2}a. This TI is a hand-made drawing inspired by Strebelle's TI \citep{Strebelle2002} by \citet{Zahner2016}. For learning, we used $z_{\rm x} = 12$ and $q=1$. Consistently with setting $z_{\rm x} = 12$ in equation (\ref{sgan2}), the training batch for each epoch was constructed by randomly cutting 64 patches of dimensions $353 \times 353$ from the TI. To generate realizations we set $z_{\rm x} = 20$ and $q=1$, thereby leading to a 400-dimensional $\textbf{Z}$ and realizations of size $609 \times 609$. The learned SGAN model deemed to work best was achieved at training epoch 25 while networks learned after epochs 25-30 started to degrade (not shown). Moreover, the produced continuous model realizations (in $\left[0,1\right]$) were processed through a median filter with kernel size of $\left(3,3\right)$ before being thresholded at the 0.5 level.

\begin{figure}[H]
\noindent\hspace{0cm}\includegraphics[width=35pc]{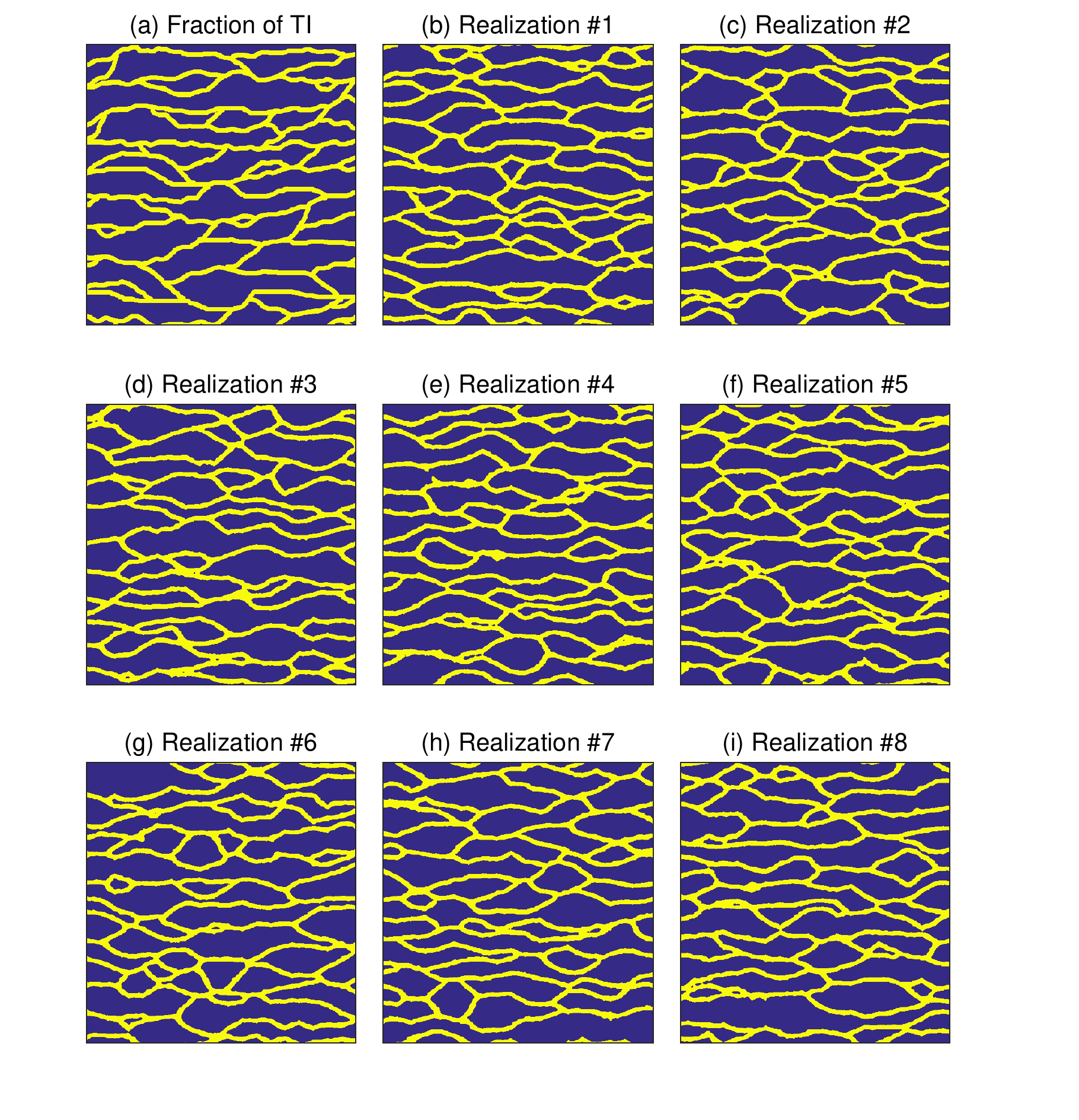}
\caption{(a) Fraction of size $609 \times 609$ of the TI shown in Figure \ref{fig2}a and (b) - (i) randomly chosen $609 \times 609$ realizations derived by our SGAN. Each realization is generated by sampling 400 random numbers from a uniform distribution, $U\left(-1,1\right)$.}
\label{fig3}
\end{figure}

\begin{figure}[H]
\noindent\hspace{0cm}\includegraphics[width=35pc]{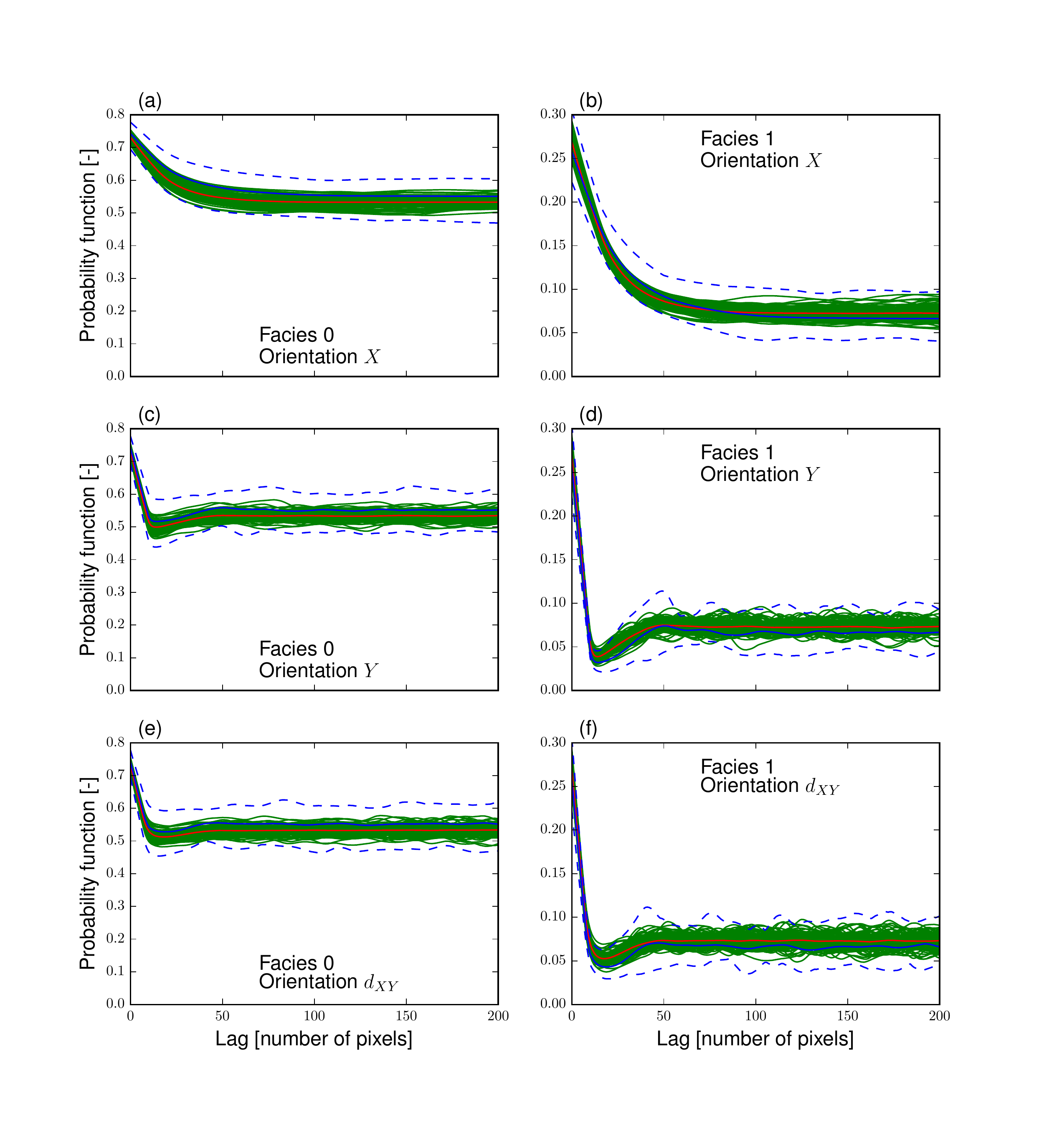}
\caption{Probability function (PF) for the 2D case study involving the binary channelized aquifer TI depicted in Figure \ref{fig2}a. The blue lines refer to 100 randomly selected patches of size $609 \times 609$ from the $2500 \times 2500$ TI with the solid blue line indicating the mean and the 2 dashed lines representing the minimum and maximum values at each lag. The green solid lines represent 100 SGAN realizations of size $609 \times 609$. The PF is calculated for each facies along spatial directions. The $x$ and $y$ symbols signify the $x$ and $y$ axes, and $d_{xy}$ represents the diagonal direction formed by the 45$\degree$ angle between the $x$ and $y$ axes.} 
\label{fig4}
\end{figure}

\begin{figure}[H]
\noindent\hspace{0cm}\includegraphics[width=35pc]{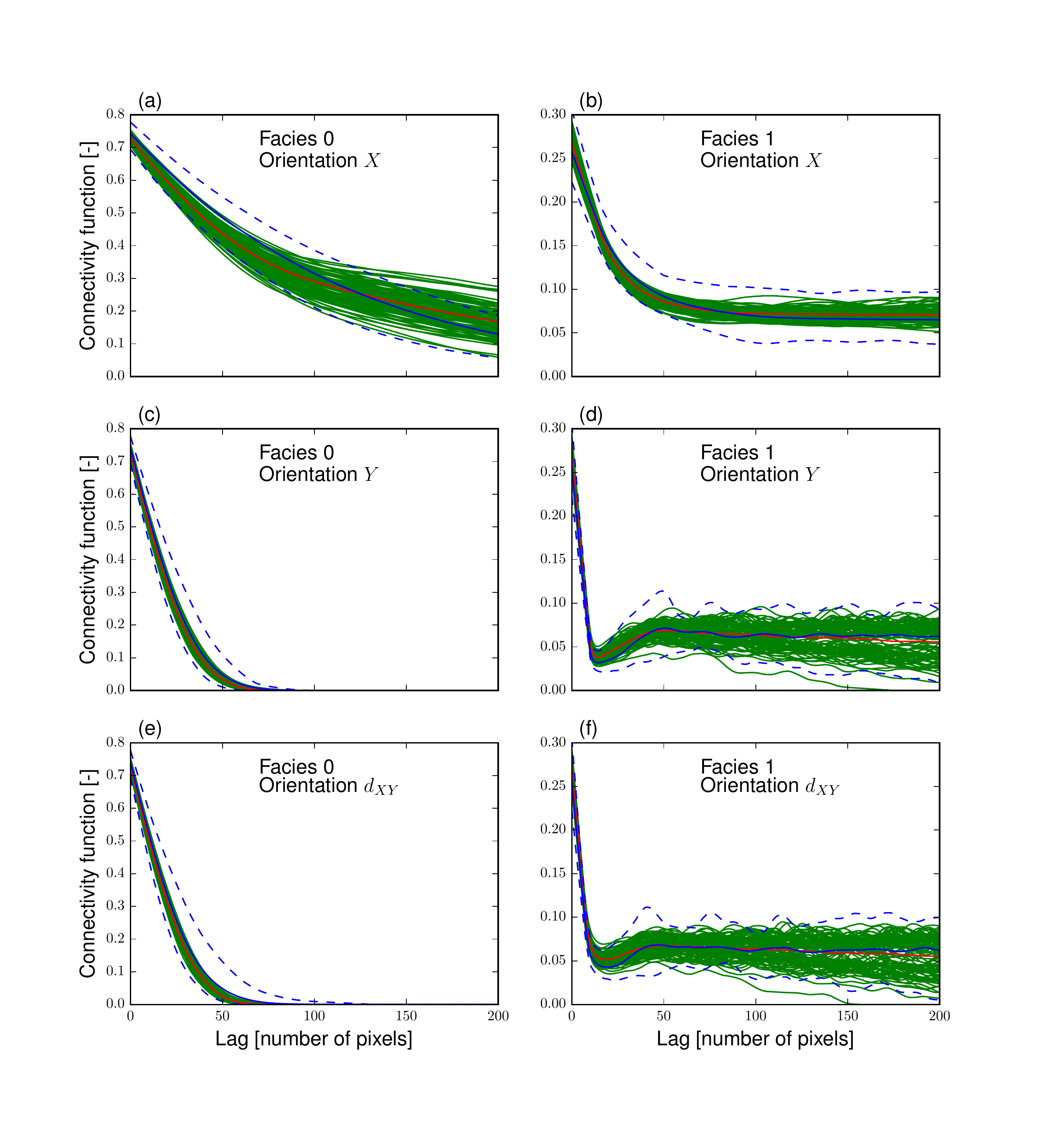}
\caption{Cluster or connectivity function (CF) for the 2D case study involving the binary channelized aquifer TI depicted in Figure \ref{fig2}a. The blue lines refer to 100 randomly selected patches of size $609 \times 609$ from the $2500 \times 2500$ TI with the solid blue line indicating the mean and the 2 dashed lines representing the minimum and maximum values at each lag. The green solid lines represent 100 SGAN realizations of size $609 \times 609$. The CF is calculated for each facies along spatial directions. The $x$ and $y$ symbols signify the $x$ and $y$ axes, and $d_{xy}$ represents the diagonal direction formed by the 45$\degree$ angle between the $x$ and $y$ axes.}
\label{fig5}
\end{figure}

Figure \ref{fig3} presents 8 randomly chosen realizations, together with a $609 \times 609$ fraction of the TI. Visually, the realizations appear consistent with the TI. Figures \ref{fig4} and \ref{fig5}  show the associated PF and CF metrics. Although some deviations exist, there is a generally good match between the realizations and the TI statistics. Also, the different realizations are slightly less variable than the TI's patches. It is important to note, however, that the ratio of the used size of a TI's patch to the size of the TI is about 1/17. As opposed to the 100 SGAN-based realizations and associated metrics, the 100 TI's patches and associated metrics are therefore not independent. With respect to facies fractions, the TI has a matrix pixel proportion of 0.74 while the corresponding average over the 100 realizations is 0.73.

Our second 2D case study involves the $400 \times 400$ tri-categorical TI shown in Figure \ref{fig2}b. This TI represents a braided river aquifer \citep{Pirot2015}. After some trial and error testing, here we used $z_{\rm x} = 5$ and $q=3$ for learning and $z_{\rm x} = 10$ and $q=3$ to generate the realizations. The size of the 64 training patches belonging to each epoch was thus $129 \times 129$, while a 300-dimensional $\textbf{Z}$ was used to produce the analyzed $289 \times 289$ realizations. For this case study only, the number of training epochs was raised from 50 to 100 because image generation quality kept improving after epoch 50. The optimally-trained network model was judged to be that produced at epoch 92. The generated realizations were not filtered but thresholded as $\leq 0.33$ = facies 0, $0.33 < \cdots \leq 0.67$ = facies 1 and $> 0.67$ = facies 2.

\begin{figure}[H]
\noindent\hspace{0cm}\includegraphics[width=35pc]{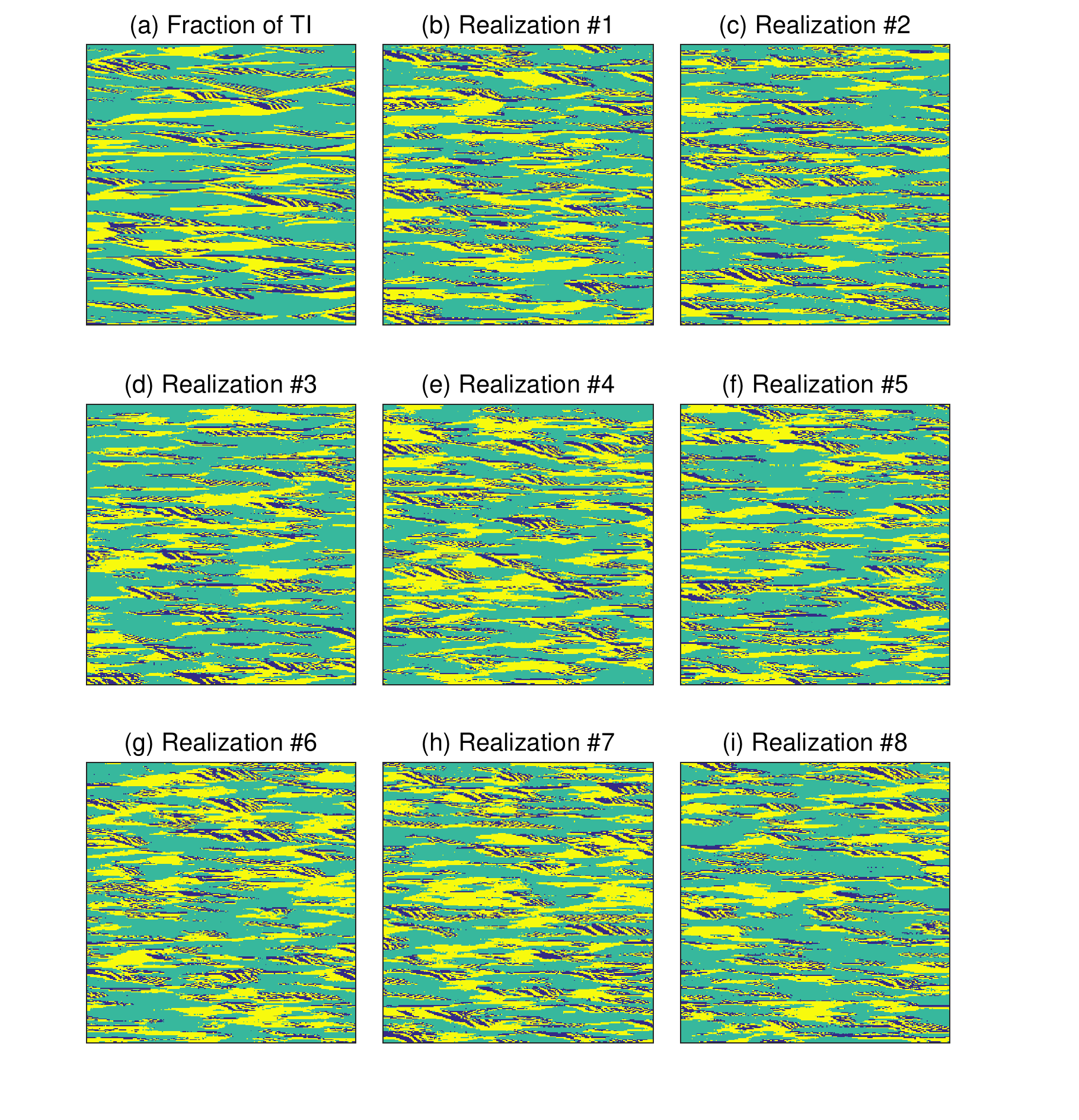}
\caption{(a) Fraction of size $289 \times 289$ of the TI shown in Figure \ref{fig2}b and (b) - (i) randomly chosen $289 \times 289$ realizations derived by our SGAN. Each realization is generated by sampling 300 random numbers from a uniform distribution, $U\left(-1,1\right)$.}
\label{fig6}
\end{figure}

\begin{figure}[H]
\noindent\hspace{-1cm}\includegraphics[width=40pc]{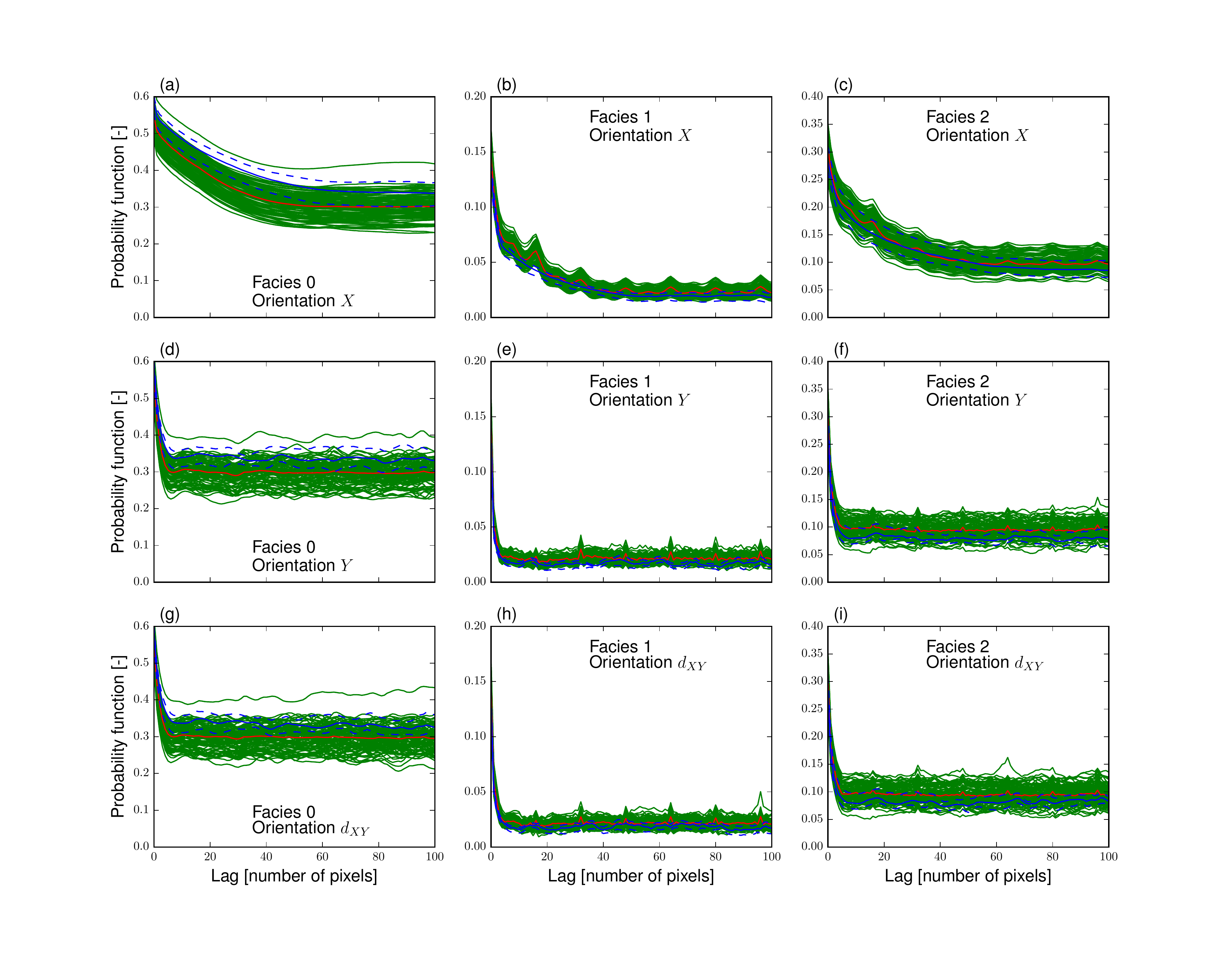}
\caption{Probability function (PF) for the 2D case study involving the tri-categorical braided river aquifer TI depicted in Figure \ref{fig2}b. The blue lines refer to 100 randomly selected patches of size $289 \times 289$ from the $400 \times 400$ TI with the solid blue line indicating the mean and the 2 dashed lines representing the minimum and maximum values at each lag. The green solid lines represent 100 SGAN realizations of size $289 \times 289$. The PF is calculated for each facies along spatial directions. The $x$ and $y$ symbols signify the $x$ and $y$ axes, and $d_{xy}$ represents the diagonal direction formed by the 45$\degree$ angle between the $x$ and $y$ axes.}
\label{fig7}
\end{figure}

\begin{figure}[H]
\noindent\hspace{-1cm}\includegraphics[width=40pc]{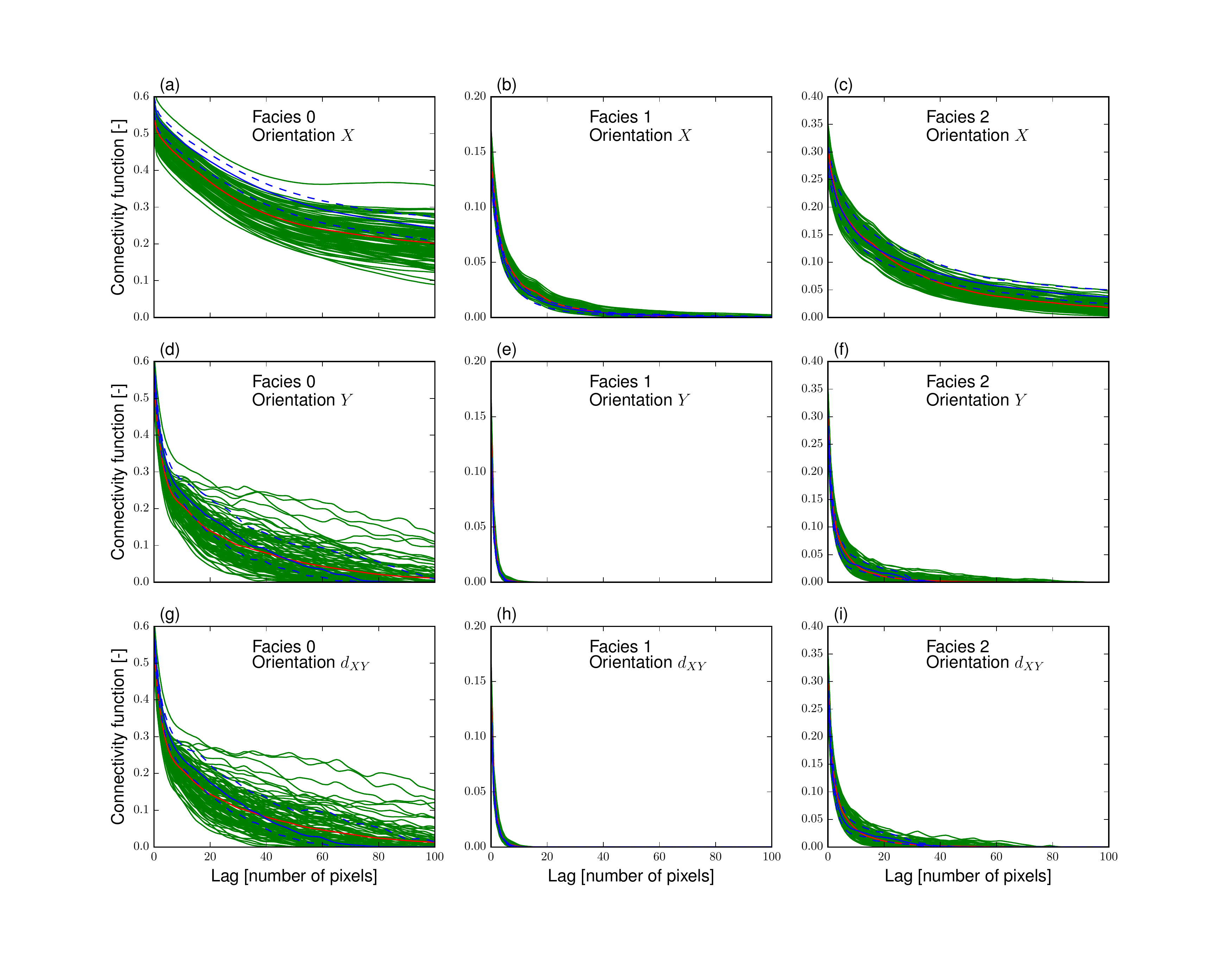}
\caption{Cluster or connectivity function (CF) for the 2D case study involving the tri-categorical braided river aquifer TI depicted in Figure \ref{fig2}b. The blue lines refer to 100 randomly selected patches of size $289 \times 289$ from the $400 \times 400$ TI with the solid blue line indicating the mean and the 2 dashed lines representing the minimum and maximum values at each lag. The green solid lines represent 100 SGAN realizations of size $289 \times 289$. The CF is calculated for each facies along spatial directions. The $x$ and $y$ symbols signify the $x$ and $y$ axes, and $d_{xy}$ represents the diagonal direction formed by the 45$\degree$ angle between the $x$ and $y$ axes.}
\label{fig8}
\end{figure}

Figure \ref{fig6} depicts 8 randomly chosen realizations, together with a $289 \times 289$ fraction of the TI. The realizations look fairly similar to the TI. Figures \ref{fig7} and \ref{fig8} show the associated PF and CF metrics. A relatively good correspondence between the TI and the realizations is observed. The main discrepancies are a larger spread in the realizations' statistics and the regular occurrence of small peaks in the PF of facies 1 and, to a lesser extent, for facies 2 (Figure \ref{fig7}). The larger spread in the SGAN-based realizations is however not surprising given that the TI's patches are strongly dependent for this example. Indeed, the ratio of the size of a TI's patch to the size of the TI is as large as (approximately) 1/2. With respect to the peaks, they are probably caused by the complex interaction between the filter size length (here 3), $dp$ and other convolutional settings (e.g., stride and padding) and the use of a smaller domain for learning than for generation. In addition, the facies fractions in the TI are moderately well reproduced by the average over 100 realizations: facies 0: 0.596 versus 0.546, facies 1: 0.119 versus 0.147,  and facies 2: 0.285 versus 0.307.

\begin{figure}[H]
\noindent\hspace{0cm}\includegraphics[width=35pc]{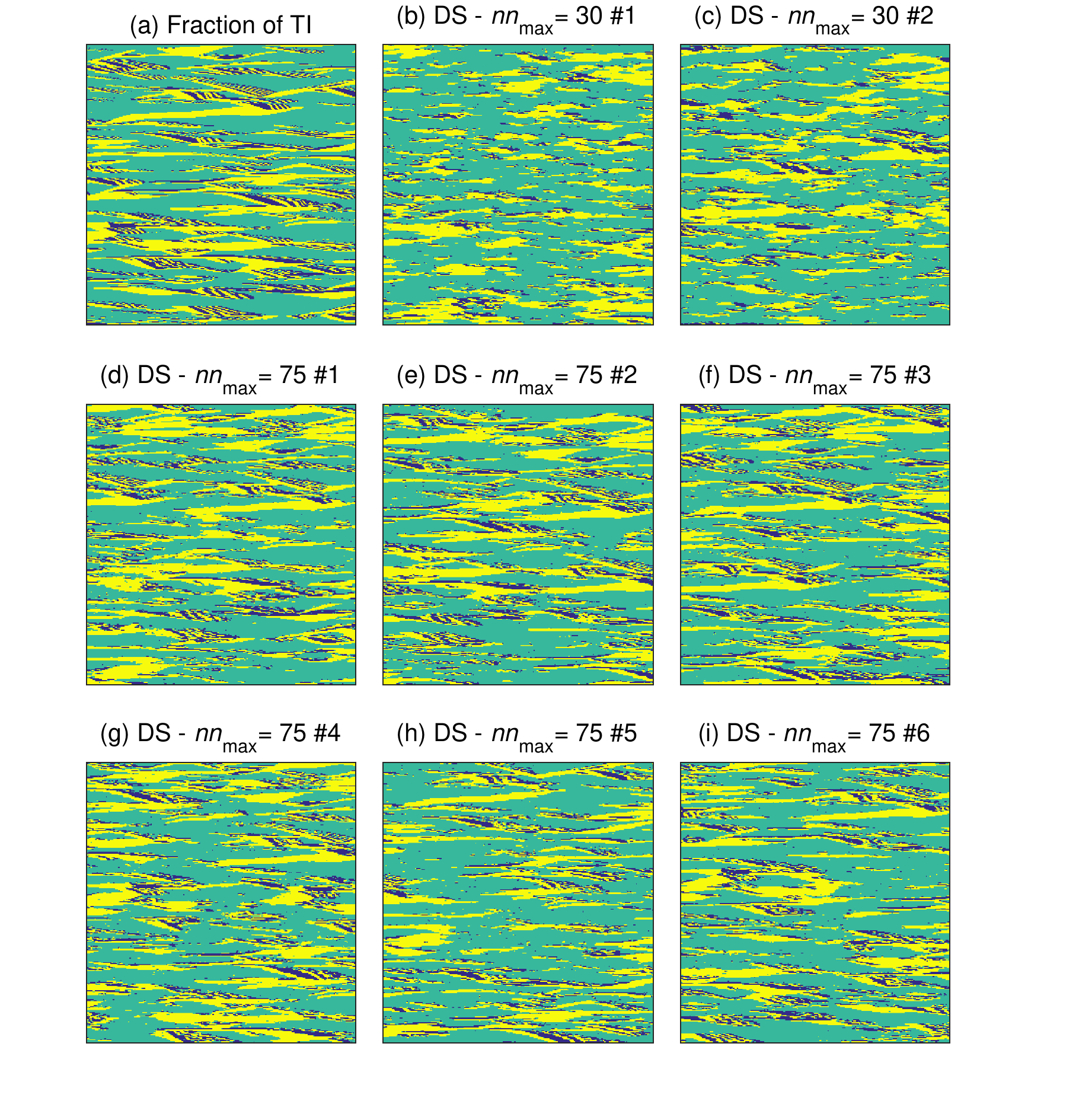}
\caption{(a) Fraction of size $289 \times 289$ of the TI shown in Figure \ref{fig2}b, and (b) - (c) randomly chosen $289 \times 289$ realizations derived by the DS algorithm with a maximum number of neighboring nodes ($nn_{\rm max}$ of 30, and (d)-(i) randomly chosen $289 \times 289$ realizations derived by the DS algorithm with $nn_{\rm max}$ = 75. On the used machine, the CPU-time per DS realization was 417 s for $nn_{\rm MAX}$ = 30 and 1180 s for $nn_{\rm max}$ = 75.}
\label{fig9}
\end{figure}

\begin{figure}[H]
\noindent\hspace{-1cm}\includegraphics[width=40pc]{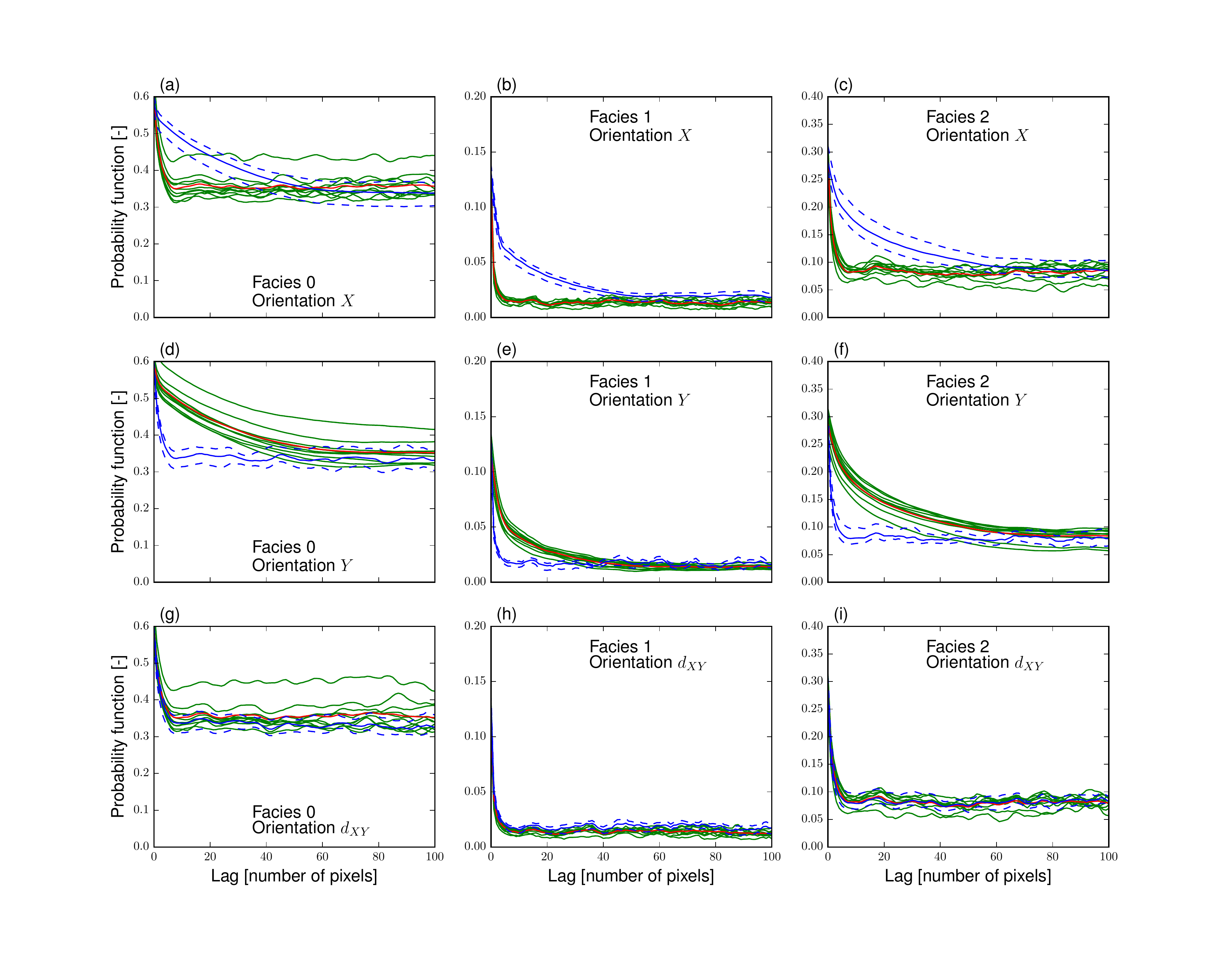}
\caption{Probability function (PF) obtained when running the DS algorithm for the 2D case study involving the tri-categorical braided river aquifer TI depicted in Figure \ref{fig2}b. The blue lines refer to 100 randomly selected patches of size $289 \times 289$ from the $400 \times 400$ TI with the solid blue line indicating the mean and the 2 dashed lines representing the minimum and maximum values at each lag. The green solid lines represent 10 realizations of size $289 \times 289$ derived by DS with the $nn_{\rm max}$ parameter set to 75. Only 10 realizations were generated given the large CPU-time per realization inherent to DS for this case study. The PF is calculated for each facies along spatial directions. The $x$ and $y$ symbols signify the $x$ and $y$ axes, and $d_{xy}$ represents the diagonal direction formed by the 45$\degree$ angle between the $x$ and $y$ axes.}
\label{fig10}
\end{figure}

\begin{figure}[H]
\noindent\hspace{-1cm}\includegraphics[width=40pc]{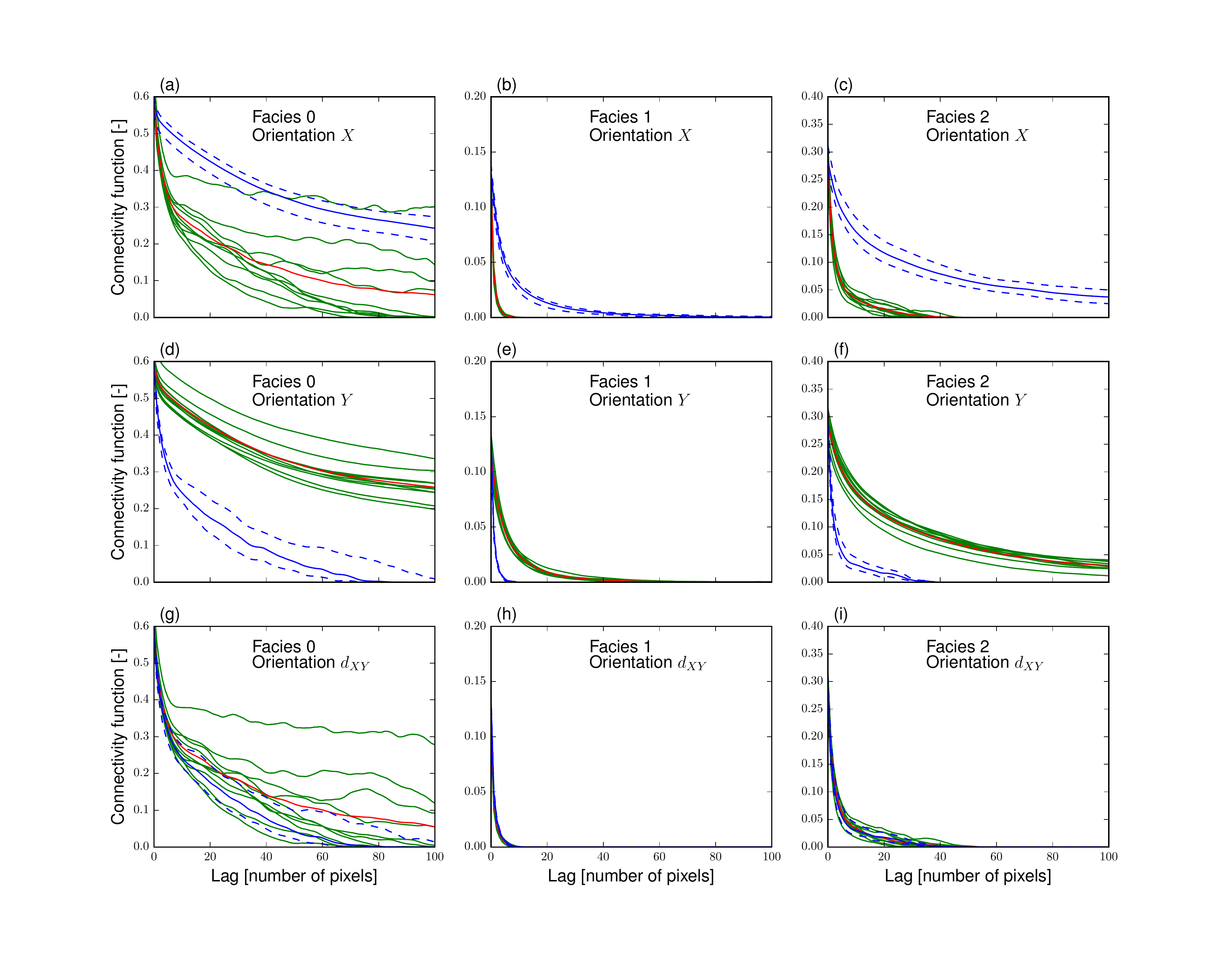}
\caption{Cluster or connectivity function (CF) obtained when running the DS algorithm for the 2D case study involving the tri-categorical braided river aquifer TI depicted in Figure \ref{fig2}b. The blue lines refer to 100 randomly selected patches of size $289 \times 289$ from the $400 \times 400$ TI with the solid blue line indicating the mean and the 2 dashed lines representing the minimum and maximum values at each lag. The green solid lines represent 10 realizations of size $289 \times 289$ derived by DS with the $nn_{\rm max}$ parameter set to 75. Only 10 realizations were generated given the large CPU-time per realization inherent to DS for this case study. The CF is calculated for each facies along spatial directions. The $x$ and $y$ symbols signify the $x$ and $y$ axes, and $d_{xy}$ represents the diagonal direction formed by the 45$\degree$ angle between the $x$ and $y$ axes.}
\label{fig11}
\end{figure}

For this case study, we performed a comparison between our trained SGAN and the well-established DeeSse (DS) MPS algorithm \citep{Mariethoz2010}. Using the standard setting of a maximum number of neighboring nodes, $nn_{\rm max}$, of 30, together with a large search radius of 120 pixels and a distance threshold of 0.01 \citep[see][for details about these algorithmic parameters]{Mariethoz2010}, producing one model realization incurs a computational cost of 417 s on the used machine. As shown in Figures \ref{fig9}b and \ref{fig9}c, the resulting realizations strongly deviate from the TI. Increasing $nn_{\rm max}$ to 75 leads to a CPU-time per realization of 1180 s. The corresponding realizations now look more similar to the TI (Figure \ref{fig9}d-i) and their facies proportions are close to those of the TI: facies 0: 0.594, facies 1: 0.118 and facies 2: 0.288. Yet the associated PF and CF statistics generally do not match those of the TI (Figures \ref{fig10} and \ref{fig11}). In contrast, after training our SGAN incurs a CPU-time per realization of only 0.1 s on the same machine, which is thus more than 10,000 times less than for DS. Furthermore, despite the unwanted peaks described above our SGAN produces more consistent PF and CF statistics (compare Figures \ref{fig7} and \ref{fig8} against Figures \ref{fig10} and \ref{fig11}, respectively).

\subsection{3D Models}
\label{3d_gen}

Our 3D case study considers the $180 \times 150 \times 120$ categorical fold TI depicted in Figures \ref{fig2}c-d (and available at \url{http://www.trainingimages.org/training-images-library.html}). Here we used $z_{\rm x} = 4$ and $q=1$ for training, together with a training batch size of 25. This implies using twenty-five $97 \times 97 \times 97$ random patches to build the training set associated with each epoch. In contrast, $z_{\rm x} = 5$ and $q=1$ were chosen to generate the realizations. This signifies that $129 \times 129 \times 129$ realizations are produced from a 125-dimensional, that is, $5 \times 5 \times 5$, $\textbf{Z}$ array. Furthermore, the selected 3D SGAN model was obtained at epoch 16 and median filtering with a $\left(3,3,3\right)$ kernel size was applied to the realizations before thresholding at the 0.5 level. The impact of median filtering was again very small (see supporting Figures S2d-f). Training the SGAN for 50 epochs took about 12 hours whereas producing a single $129 \times 129 \times 129$ realization with the trained network takes 12 s on the used last generation intel\textsuperscript{\textregistered} i7 CPU. Note that this time includes the (relatively slow) post-filtering and (relatively quick) post-thresholding, whereas the SGAN generation itself requires only about 5 s.

\begin{figure}[H]
\noindent\hspace{-1.0cm}\includegraphics[width=40pc]{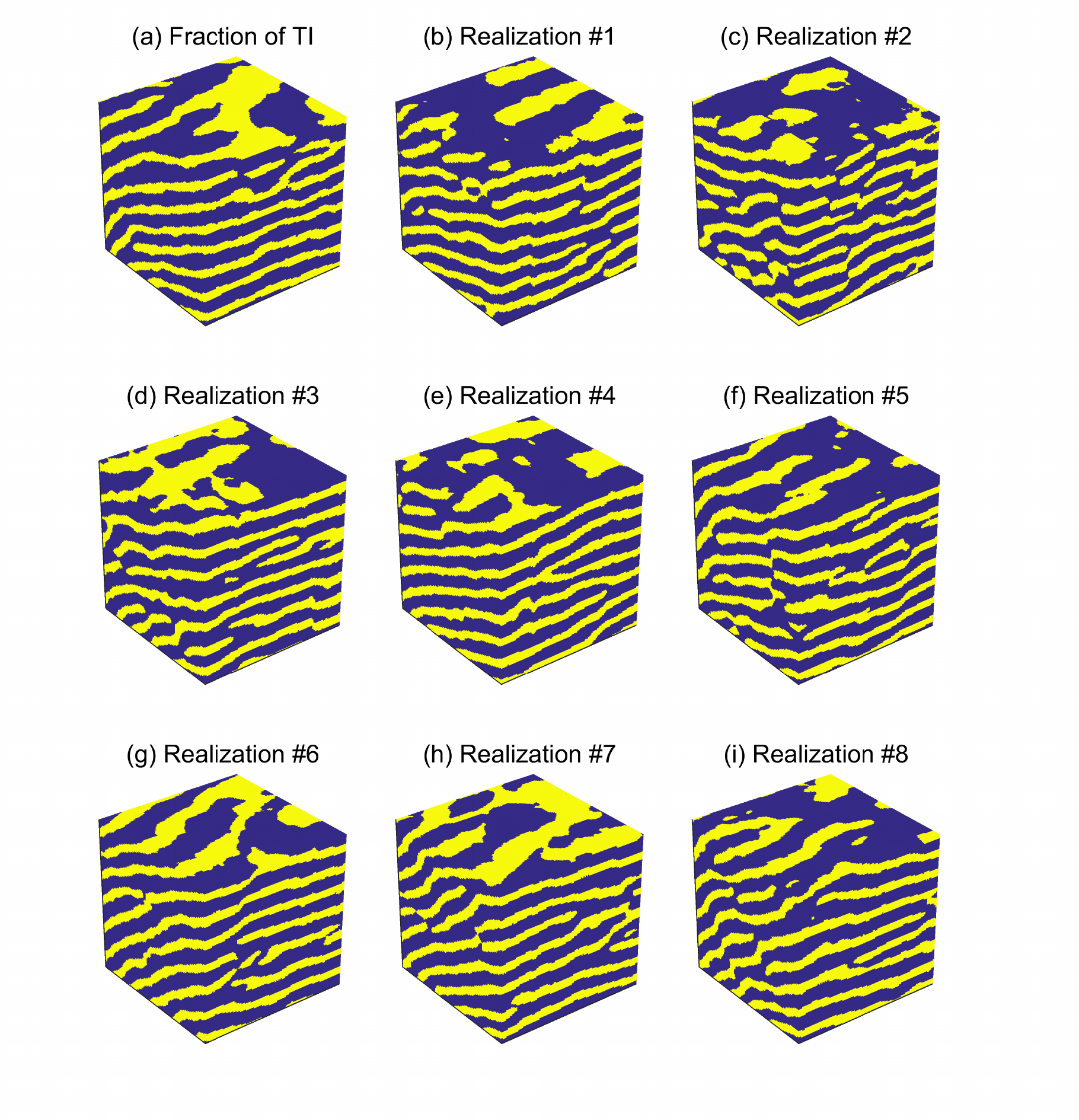}
\caption{(a) Fraction of size $120 \times 120 \times 120$ of the TI shown in Figure \ref{fig2}d and (b) - (i) randomly chosen $120 \times 120 \times 120$ realizations derived by our 3D SGAN. Each realization is generated by sampling 125 random numbers from a uniform distribution, $U\left(-1,1\right)$. The $180 \times 150 \times 120$ TI and original $129 \times 129 \times 129$ realizations (see main text for details) were all cropped to $120 \times 120 \times 120$ for visual convenience.}
\label{fig12}
\end{figure}

\begin{figure}[H]
\noindent\hspace{-1.0cm}\includegraphics[width=40pc]{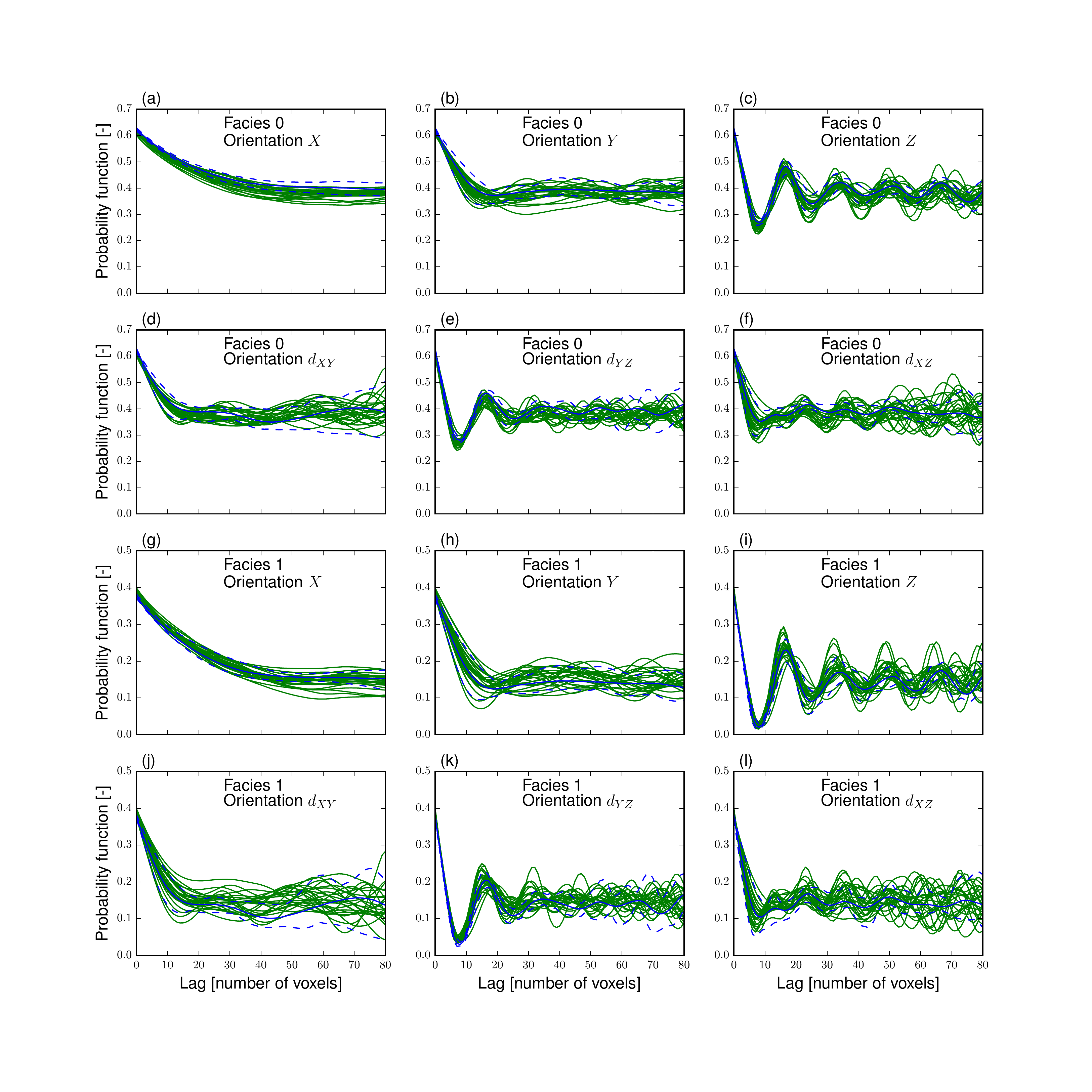}
\caption{Probability function (PF) for the 3D case study involving the binary fold aquifer TI depicted in Figure \ref{fig2}d. The blue lines refer to 25 randomly selected patches of size $120 \times 120 \times 120$ from the $180 \times 150 \times 120$ TI with the solid blue line indicating the mean and the 2 dashed lines representing the minimum and maximum values at each lag. The green solid lines represent 25 SGAN realizations of size $120 \times 120 \times 120$. The PF is calculated for each facies along spatial directions. The $x$, $y$ and $z$ symbols signify the $x$, $y$ and $z$ axes. The $d_{xy}$, $d_{yz}$ and $d_{xz}$ symbols represent the diagonal directions formed by the 45$\degree$ angle from the $x$-axis in the $xy$ plane, from the $y$-axis in the $yz$ plane and from the $x$-axis in the $xz$ plane, respectively. Note that due to the presence of large connected features, the output of the cluster or connectivity function (CF) is almost identical to the displayed PF's output for this case study.}
\label{fig13}
\end{figure}

The TI and 8 (randomly chosen realizations) are presented in Figure \ref{fig12}, where the $180 \times 150 \times 120$ TI and every $129 \times 129 \times 129$ model realization was cropped to $120 \times 120 \times 120$ for visual convenience. The realizations show similar patterns as the TI despite a slight over-representation of broken channels and the rare occurrence of small isolated patches. Also, the PF (Figure \ref{fig13}) and CF (see supporting Figure S3) statistics of the TI are relatively well matched by the 25 realizations. Here we do not show the CF statistics as they are almost identical to the depicted PF statistics. This is caused by the presence of large connected features. Basically, if two voxels separated by a given lag belong to the same facies (as expressed by the PF), then it is extremely likely that they are connected (as expressed by the CF). In addition, no proper comparison can be made between the respective spreads of the realizations and the TI'patches as these TI's patches are highly dependent. Indeed, the ratio of the size of a TI's patch to the size of the TI is about 1/2 here. This large dependency might therefore cause the smaller spread observed for the TI's patches compared to the SGAN realizations. As of facies fractions, the fraction of matrix voxels is 0.62 for the TI against 0.61 in average over the 25 realizations. 

\section{Inverse Problems}
\label{results_inv}
\subsection{Case Study 1: 2D Steady-State Flow}
\label{inv2d}

\begin{figure}[H]
\noindent\hspace{-1.5cm}\includegraphics[width=42pc]{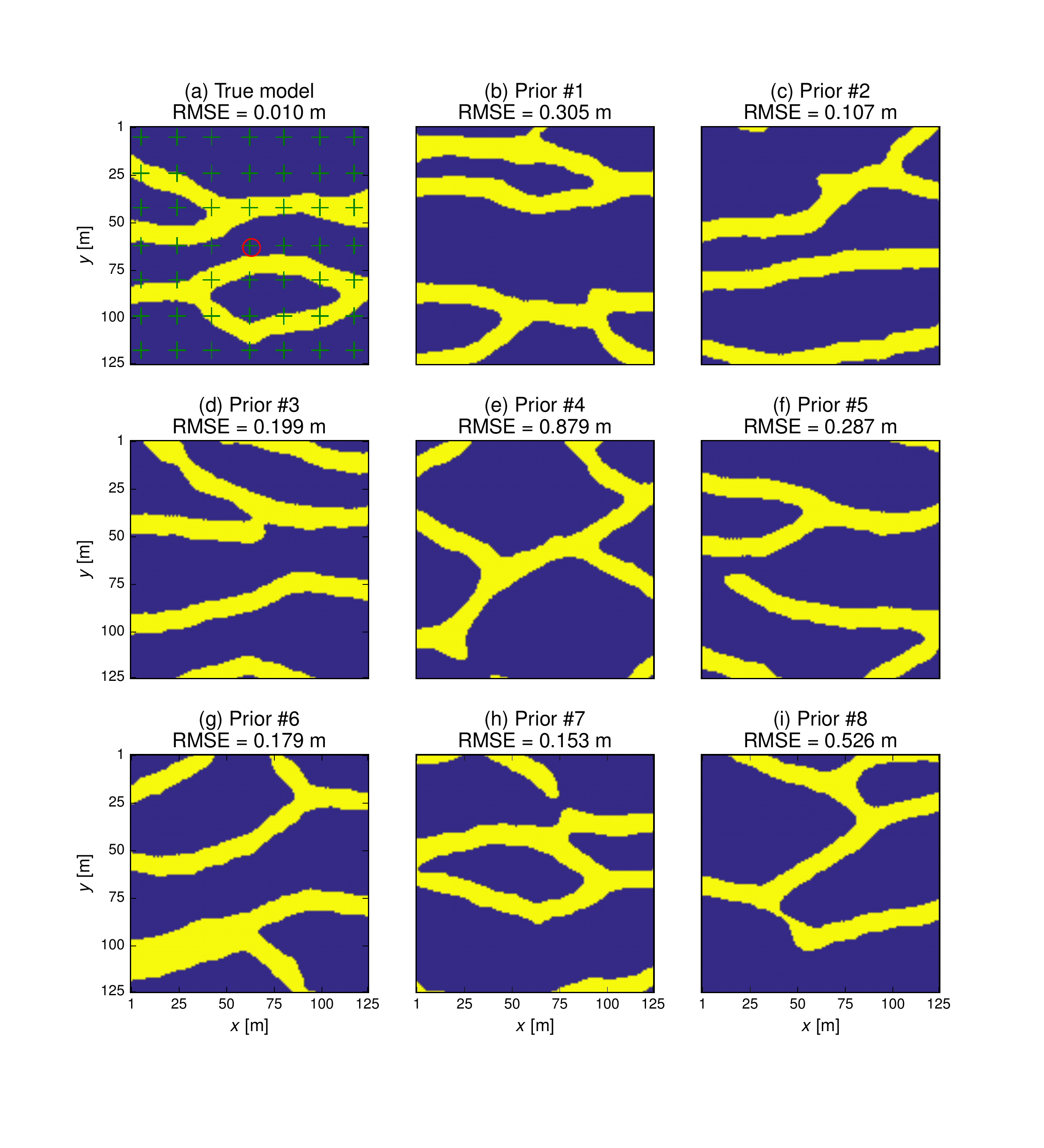}
\caption{(a) True model and (b-i) eight randomly chosen prior realizations. The red circle and green crosses in subfigure (a) mark the location of the pumping well and the piezometers, respectively. The models' dimensions are $125 \times 125$.}
\label{fig14}
\end{figure}

Our first inversion case study considers 2D steady-state flow within a channelized aquifer that is consistent with the TI shown in Figure \ref{fig2}a. The total number of pixels is $125 \times 125 = 15,625$, and the 125 $\times$ 125 aquifer domain lies in the $x-y$ plane with a grid cell size of 1 m and a thickness of 1 m. Channel and matrix materials (see Figure \ref{fig14}a and supporting Figure S4) are assigned hydraulic conductivity values of 1 $\times$ 10$^{-2}$ m/s and 1 $\times$ 10$^{-4}$ m/s, respectively. MODFLOW 2005 \citep{Harbaugh2005} is used to simulate steady state groundwater flow with no-flow boundaries at the upper and lower sides and a lateral head gradient of 0.01 (-) with water flowing in the $x$-direction. Water is extracted at a rate of 0.001 m$^3$/s by a well located at the center of the domain. The measurement data were formed by acquiring simulated heads at 49 locations that are regularly spread over the domain (Figure \ref{fig14}a and supporting Figure S4). A Gaussian white noise with standard deviation of 0.01 m was then used to corrupt these data. For the selected white noise realization, the measurement data have a root-mean-square-error (RMSE) of 0.0096 m. The corresponding signal-to-noise-ratio (SNR), defined as the ratio of the average RMSE obtained by drawing prior realizations with our SGAN algorithm to the noise level is as large as 44. Figures \ref{fig14}b-i depict 8 (randomly chosen) prior realizations that demonstrate large prior variability.
The reference model (Figure \ref{fig14}a) was generated using our SGAN. This was achieved by randomly drawing a 25-dimensional ($5 \times 5$) $\bm{\uptheta} = \textbf{Z} \sim U\left(-\textbf{1},\textbf{1}\right)$, which results in a $129 \times 129$ model realization (see equation (\ref{sgan2})). The obtained realization was cropped to $125 \times 125$ to form the reference model while the models generated during the MCMC sampling were similarly cropped. The DREAM$_{\rm \left(ZS\right)}$ sampler was ran in parallel, using 8 interacting Markov chains distributed over 8 CPUs. Uniform priors in the $\left[-1,1\right]$ range were selected for the 25-dimensional $\bm{\uptheta}$. 

\begin{figure}[H]
\noindent\hspace{-1.5cm}\includegraphics[width=42pc]{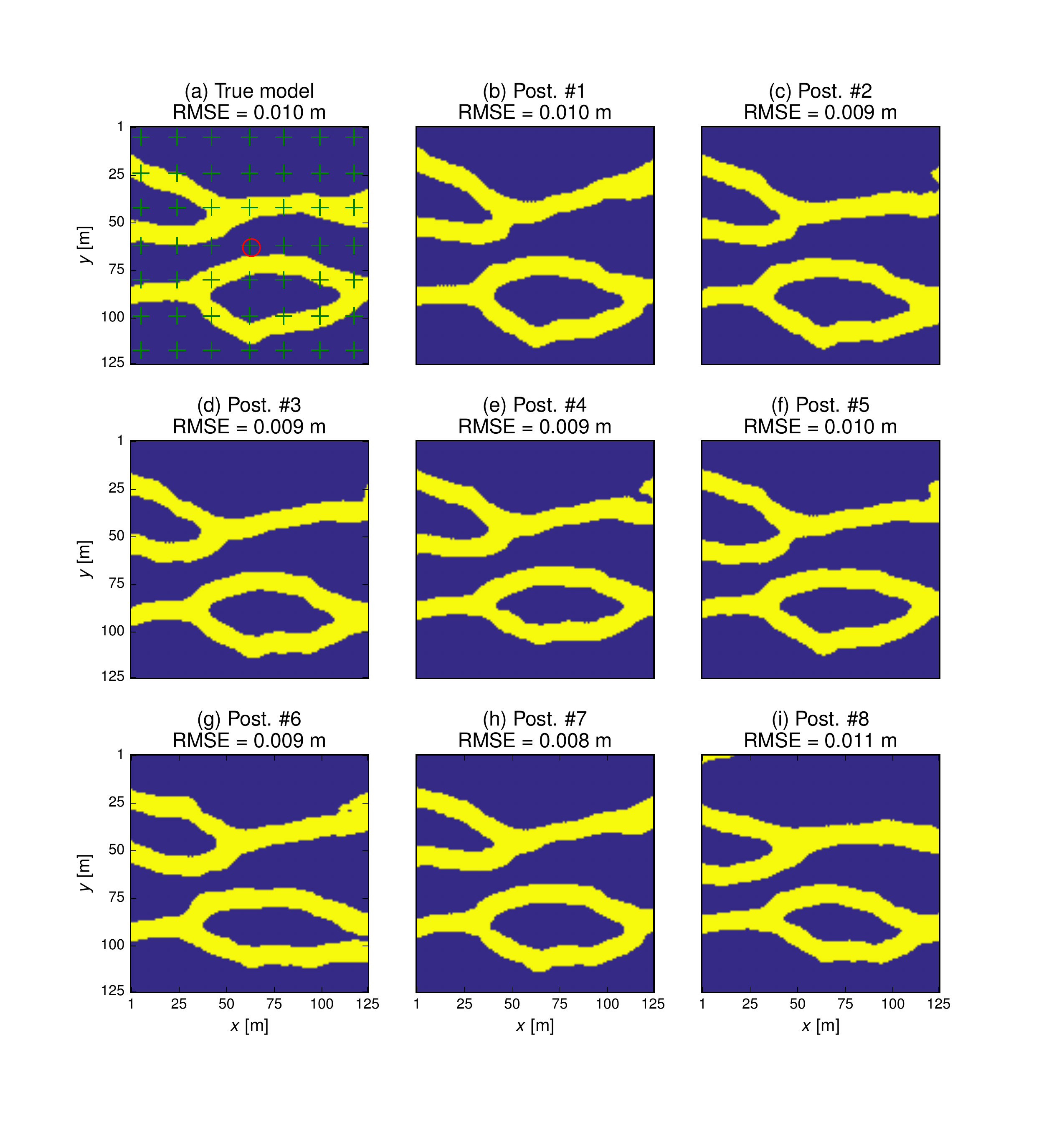}
\caption{(a) True model and (b-i) eight randomly chosen posterior realizations derived by DREAM$_{\rm \left(ZS\right)}$. The red circle and green crosses in subfigure (a) mark the location of the pumping well and the piezometers, respectively. The models' dimensions are $125 \times 125$.}
\label{fig15}
\end{figure}

The chains start to jointly sample the posterior distribution, $p\left(\bm{\uptheta} | \textbf{d}  \right)$, after a (serial) total of approximately 96,000 iterations, that is, 12,000 parallel iterations per chain (not shown). The sampled realizations closely resemble the true model and the posterior variability is rather small (Figure \ref{fig15}). After a total of 387,200 MCMC iterations, that is, 48,400 iterations in each chain, the \citet{Gelman-Rubin1992} convergence diagnostic, $\hat{R}$, is satisfied (i.e., $\hat{R} \leq 1.2$) for every sampled parameter \citep[see, e.g.,][for details about the use of $\hat{R}$ with DREAM$_{\rm \left(ZS\right)}$]{Laloy2015}.

To evaluate the inclusion of direct conditioning data, $\textbf{x}_m$, within the inversion, we performed a second MCMC run where the facies of the 49 grid cells where the wells are located (green crosses in Figure \ref{fig14}a) are known and incorporated in equations (\ref{mcmc3}) and (\ref{mcmc4}). This requires setting the $\sigma_x$ term in equation (\ref{mcmc3}) which weights $\ell\left(\bm{\uptheta} | \textbf{d}  \right)$ and $\ell\left(\bm{\uptheta} | \textbf{x}_m  \right)$ against each other in equation (\ref{mcmc4}). Within a pure Bayesian framework, data weighting should only be based on the standard deviations of the data errors. Yet here $\sigma_x$ is not known a priori. Limited testing with $\sigma_x = 1$, $\sigma_x = 0.5$ and $\sigma_x = 0.1$ revealed that for the considered case study, $\sigma_x = 1$ is sub-optimal with respect to honoring $\textbf{x}_m$ while $\sigma_x = 0.1$ induces slightly too large (indirect hydrologic) data misfits. In contrast, the choice of $\sigma_x = 0.5$ offers a good tradeoff (see below) and was therefore selected for our calculations.

The inclusion of direct conditioning data is found to be beneficial with respect to convergence of the MCMC sampling. Indeed, the 8 chains now start to jointly sample the posterior distribution, $p\left(\bm{\uptheta} | \textbf{d}, \textbf{x}_m  \right)$, after 10,000 parallel iterations per chain (against 12,000 for the unconditional case) and, more important, the $\hat{R}$ convergence diagnostic is satisfied for every sampled parameter after 35,300 iterations in each chain (against 48,400 for the unconditional case). With respect to conditioning accuracy, across the last 160 posterior realizations 88\% of the realizations honor all of the 49 conditioning data (100\% conditioning accuracy), and 100\% of the realizations present at most 1 mismatching datum (98\% conditioning accuracy). Similar statistics for the previous case that did not consider direct conditioning data (Figure \ref{fig15}) are 0\% of the posterior realizations (across the last 160 ones) honor all of the 49 ``true" facies or present at most one mismatching ``true" facies, while 100\% of the posterior realizations present at most 8 mismatching facies compared to the ``true" model (84\% conditioning accuracy). This suggests that our strategy to account for direct conditioning data within the inversion works rather well.

\subsection{Case Study 2: 3D Transient Hydraulic Tomography}
\label{inv3d}

\begin{figure}[H]
\noindent\hspace{-1.25cm}\includegraphics[width=42pc]{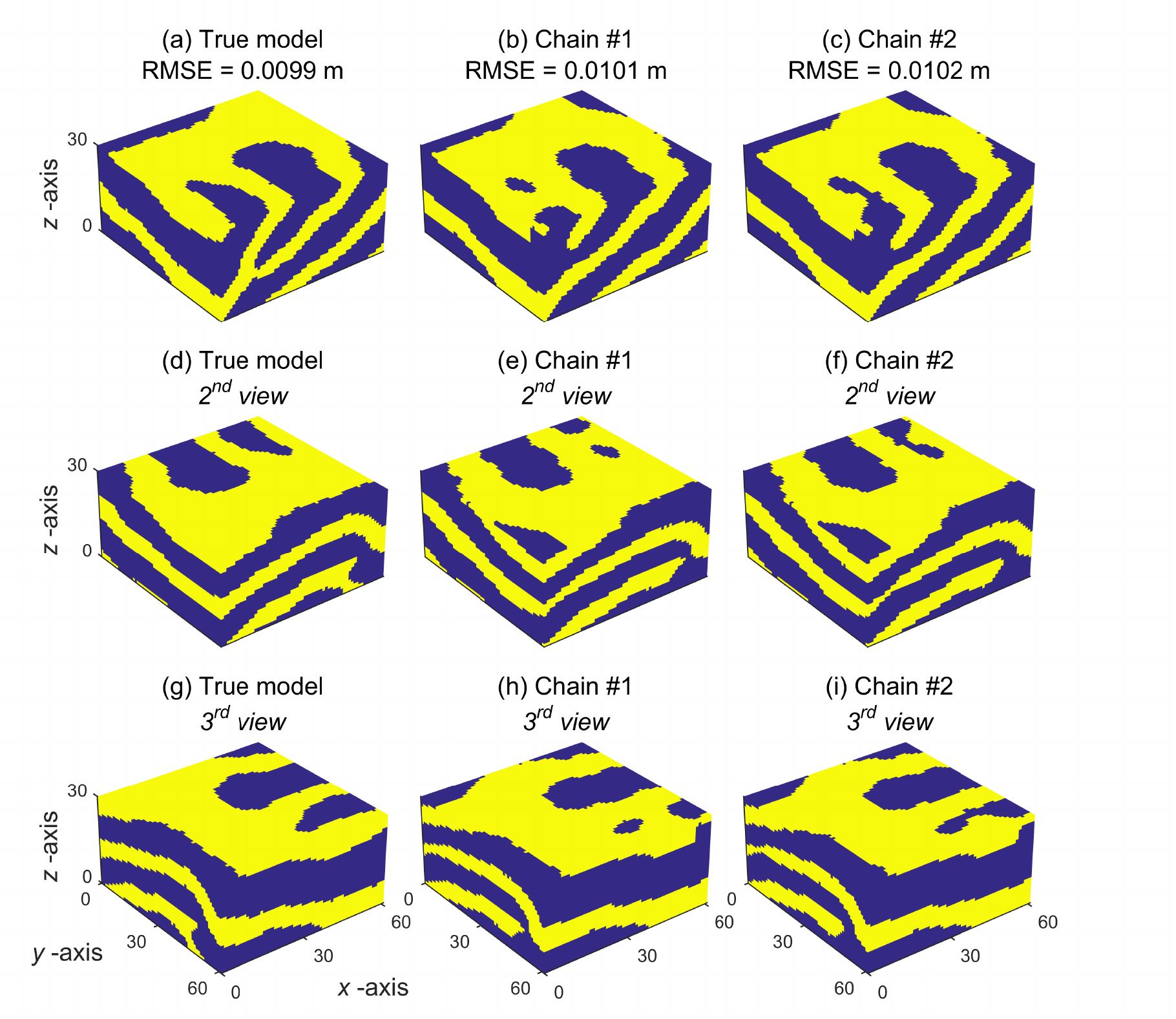}
\caption{a) True model and (b-i) states of the 8 Markov chains evolved by DREAM$_{\rm \left(ZS\right)}$ after 50,000 iterations per chain for the inverse case study 2 (see section \ref{inv3d}). The models' dimensions are $30 \times 61 \times 61$.}
\label{fig16}
\end{figure}

Our last case study consists of a 3D transient hydraulic tomography example \citep[e.g.,][]{Cardiff2013}. Transient head variations induced by discrete multilevel pumping tests were simulated using MODFLOW2005 within a 3D confined aquifer of size 30 $\times$ 61 $\times$ 61 with a voxel size of 1 m $\times$ 1 m $\times$ 1 m. The total number of voxel values is thus $30 \times 61 \times 61 = 111,630$. The considered aquifer honors the categorical fold TI presented in Figures \ref{fig2}c-d. The true model is depicted in Figure \ref{fig16}a. It was obtained by feeding our 3D SGAN with a randomly drawn 27-dimensional ($3 \times 3 \times 3$) $\bm{\uptheta} = \textbf{Z} \sim U\left(-\textbf{1},\textbf{1}\right)$. This creates models of size $65 \times 65 \times 65$ (equation (\ref{sgan2})) that are subsequently cropped to $30 \times 61 \times 61$. 

\begin{figure}[H]
\noindent\hspace{-3.5cm}\includegraphics[width=50pc]{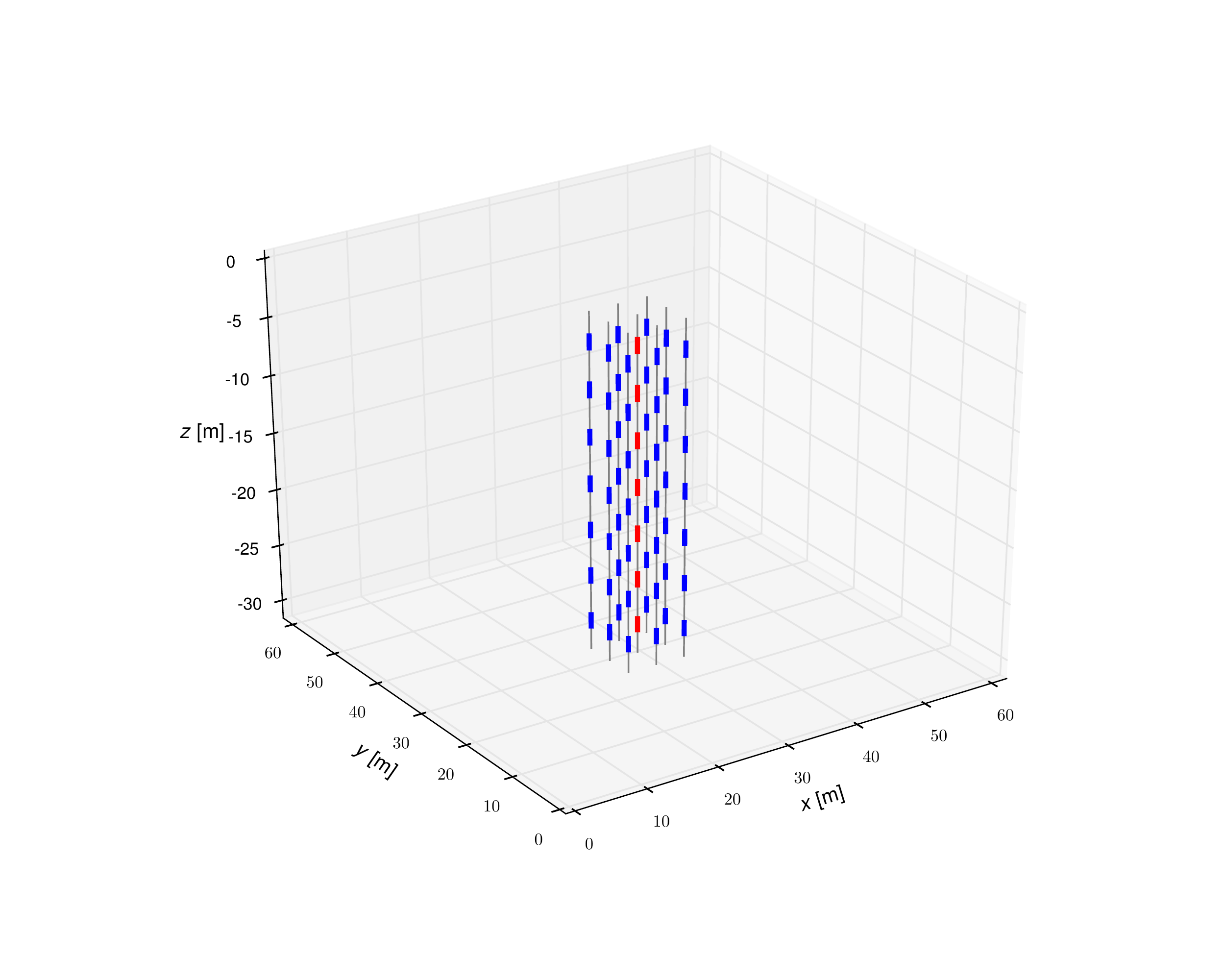}
\caption{Pumping and observation locations in the synthetic 3D hydraulic tomography setup. The vertical gray lines represent the multilevel pumping and observation wells. The red line segments in the central well are the pumping locations and the blue line segments in the other wells are the head measurement locations. The vertical direction ($z$-axis) is exaggerated by a factor 2.}
\label{fig17}
\end{figure}

Channel and matrix materials are assigned hydraulic conductivity values of 1 $\times$ 10$^{-4}$ m/s and 1 $\times$ 10$^{-6}$ m/s, respectively. The multilevel discrete pumping setup includes a 30-m deep, central multi-level well in which water is sequentially extracted every 4 m along a 1-m long screen (at depths of 3, 7, 11, 15, 19, 23 and 27 m respectively) during 30 minutes at a rate of 10 l/min. This central well is surrounded by 8 multilevel piezometers where drawdowns are recorded every 4 m along a 1-m long screen during each pumping sequence. Locations of the multilevel pumping well, the 8 multilevel observation wells and their associated screens are displayed in Figure \ref{fig17}. For each drawdown curve, data collected at the same four measurement times were selected, leading to a total of $8 \times 7 \times 7 \times 4 = 1568$ measurement data. These measurement times were considered to be the four most informative ones after visual inspection of several drawdown curves (not shown). These data were corrupted with a Gaussian white noise using again a standard deviation of 0.01 m. For the chosen white noise realization, this caused a RMSE of 0.0099 m and the associated SNR is 22. Prior variability is large, with prior models having RMSE values that are 15 to 28 times larger than that of the true model. Given the involved nonlinearities, the large number of measurement data (1568) and the relatively large SNR (22), we deem this inverse problem to be quite challenging.

For this case study, the DREAM$_{\rm \left(ZS\right)}$ sampler evolves 16 Markov chains in parallel using 16 CPUs while uniform priors in $\left[-1,1\right]$ are assumed for the 27 dimensions of $\bm{\uptheta}$. The allowed computational expense was 50,000 iterations per chain, which represents a computing time of 6.5 days on a 16-core workstation. After consumption of this budget, the 16 chains converge towards a data misfit in the range of 0.0101 m - 0.0103 m (Figures \ref{fig16}b-c, \ref{fig16}e-f, \ref{fig16}h-i). This interval is close to the target level of 0.0099 m. Nevertheless, it indicates that the exact posterior mode has not been sampled yet. The variability among the sampled models is rather small (Figures \ref{fig16}b-c, \ref{fig16}e-f, \ref{fig16}h-i). This is likely due to the combination of two factors: (1) the peakedness of the likelihood function (equation (\ref{mcmc2})) caused by the large number of measurement data with small noise, and (2) the difficulties encountered by the MCMC algorithm for exploring this complex target distribution. Nevertheless, the recovered models (Figures \ref{fig16}b-c, \ref{fig16}e-f, \ref{fig16}h-i) are visually rather close to the true model (Figure \ref{fig16}a). Also notice that the mean RMSE of the sampled models quickly drops from about 0.22 m to 0.012 m within 5000 - 10,000 iterations per chain (i.e., one day of calculation). For the remaining of the alloted computational time, the mean sampled RMSE then slowly decreases from 0.012 m to 0.0101 - 0.0103 m.

\section{Discussion}
\label{discussion}

Our results demonstrate that our proposed SGAN-based inversion approach works well. Compared to our previous work with a totally different type of neural network \citep{Laloy2017}, the proposed SGAN architecture solves three important issues: it relies on a single TI, can deal with multi-categorical TIs and builds a low-dimensional representation where each independent low-dimensional variable only influences a specific region of the generated model realization. The latter point allows us to solve inverse problems that have smaller noise and larger dimensionality compared to those considered in \citet{Laloy2017}. In addition, our proposed approach has one more advantage for inversion: it permits an even larger compression ratio. The latter can be above 10,000 for a 3D binary channelized aquifer. Yet for our synthetic 3D transient hydraulic tomography the posterior distribution was not appropriately sampled. Using more than 16 parallel DREAM$_{\rm \left(ZS\right)}$ chains and letting the MCMC sampling continue beyond 50,000 MCMC iterations per parallel chain would likely help is in this respect.

In addition to probabilistic inversion, we suggest that our 2D/3D SGAN also shows some potential for uncertainty quantification tasks that involve repeated unconditional geostatistical simulations. The main advantage of our SGAN for simulation is speed once training has been achieved, even though this training can take several hours. As stated earlier, once the SGAN is trained a 3D model realization containing 1-2 millions of grid cells can be generated in a matter of seconds. To provide more insights into the simulation efficiency or our approach when a GPU is available for training, we re-examine our comparison against the well-known DeeSse code \citep[DS,][]{Mariethoz2010} for the $400 \times 400$ braided river tri-categorical TI (depicted in Figure \ref{fig2}b). For this particular case study, our SGAN generates more consistent realizations and incurs a computational time per realization  of 0.1 s on the used CPU after a 6-hours long training (on the separately used GPU). On the same CPU, DS requires 1180 s to produce one realization. Producing 10$^2$ and 10$^3$ realizations with our SGAN thus requires $6 + 10^2 \times 0.1/3600$ $\approx$  6.0 h, and $6 + 10^3 \times 0.1/3600$ $\approx$  6.0 h as well. In contrast, producing 10$^2$ and 10$^3$ realizations with DS necessitates $10^2 \times 1180/3600 $ $\approx$ 32.8 h and $10^3 \times 1180/3600 $ $\approx$ 327.8 h. Faster MPS algorithms than DS are however available \citep[e.g.,][]{Li2016,Tahmasebi-Sahimi2016a,Tahmasebi-Sahimi2016b}. Since training time is expected to reduce a lot in the near future (and already has at the time of writing) owing to the quick evolution of deep learning software, this warrants further investigations.

An important limitation of our proposed approach is that direct conditioning to point data cannot be achieved for pure geostatistical simulation. A possible solution to this problem is to (1) train the network, (2) produce a realization, $\hat{\textbf{X}}$, from a given low-dimensional $\textbf{Z}$ vector and (3) optimize the values in $\textbf{Z}$ such that $\hat{\textbf{X}}$ honors the available conditioning point data. Limited testing in this direction shows encouraging results although the induced computational overhead needs to be reduced. Another potential problem for simulation and, to a lesser extent, inversion is training time. The herein reported 3-12 h are for a GPU Tesla K40 and training on one or more CPUs would likely take substantially longer. We expect that both (deep learning) software and (GPU) hardware development will make training significantly faster in the near future. It woud also be desirable to further improve the simulation quality of our SGAN. We expect this can be done by further optimizing the network architecture.  Furthermore, it would be highly beneficial to make training more stable such that the simulation accuracy never decreases when using more epochs. This can likely be done by using a different loss (or distance) function than that of equation (\ref{sgan3}). Indeed, recent advances suggest that various approximations of the so-called Earth-Mover (EM) or Wasserstein-1 distance improve training behavior substantially upon the classical equation (\ref{sgan3}) \citep[][]{Arjovsky2017}. Lastly, extension to rectangular rather than cubic simulation domains is straightforward.

\begin{figure}[H]
\noindent\hspace{-1.25cm}\includegraphics[width=42pc]{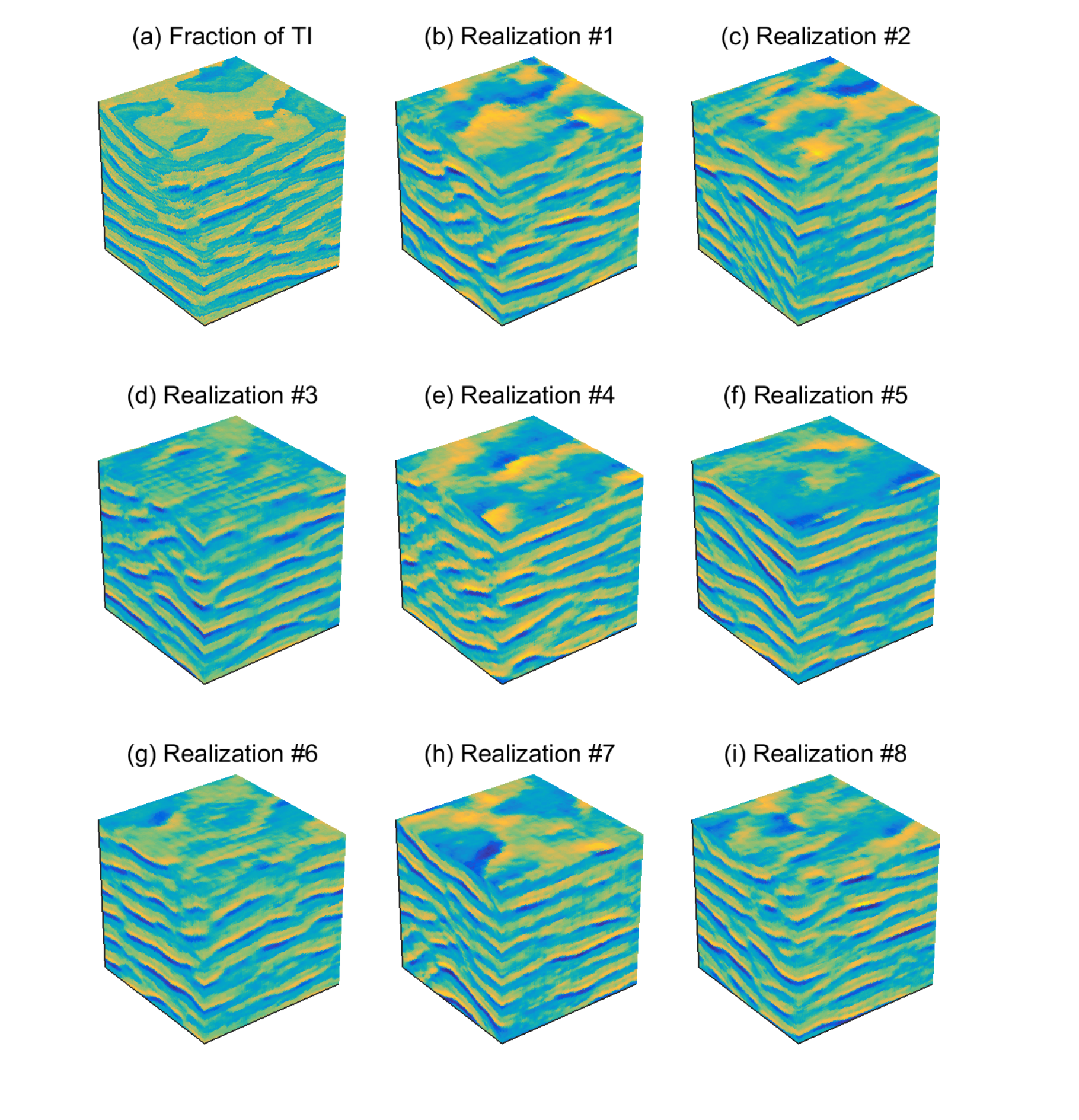}
\caption{(a) Fraction of size $120 \times 120 \times 120$ of the $180 \times 150 \times 120$ continuous fold TI (available at \protect\url{http://www.trainingimages.org/training-images-library.html}) and (b) - (i) randomly chosen $120 \times 120 \times 120$ realizations derived by our 3D SGAN. Each realization is generated by sampling 375 random numbers from a uniform distribution, $U\left(-1,1\right)$. The $180 \times 150 \times 120$ TI and original $129 \times 129 \times 129$ realizations (see main text for details) were all cropped to $120 \times 120 \times 120$ for visual convenience.}
\label{fig18}
\end{figure}

Even if not considered or demonstrated herein, we would like to stress that it might be possible to train our 2D/3D SGAN on continuous TIs as well. Figure \ref{fig18} diplays model realizations obtained by our 3D SGAN for the continuous version of the 3D categorical fold depicted in Figures \ref{fig2}c-d (a fraction of this continuous TI is shown in Figure \ref{fig18}a). These $129 \times 129 \times 129$ realizations were obtained using $z_{\rm x} = 5$ and $q=3$, that is, a 375-dimensional $\textbf{Z}$, and training took about 12 h on the used GPU. Also, a rank transformation was used to better mimick the continuous range found in the TI: the continuous values in the realizations were ranked and replaced by values with equal rank in the TI. Though far from perfect, we believe these results are promising. Lastly, it is worth noting that our approach also works with multiple (univariate) TIs \citep[such as in, e.g.,][]{Laloy2017}.

\section{Conclusion}
\label{conclusion}

We present a new training-image based inversion approach for complex geologic media that relies on a deep neural network of the generative adversarial network (GAN) kind. Our proposed spatial GAN (SGAN) can generate 2D and 3D unconditional realizations from a categorical training image (TI) and, in principle, a continuous TI as well. Compared to existing geostatistical methods, it has the advantage that it generates model realizations that capture MPS statistics of the TI using a (very) low-dimensional representation of the original model domain. This allows for efficient probabilistic inversion. For the considered case studies, training our SGAN takes 3-12 hours. Once the network is trained, generating a realization becomes very fast. For instance, a 3D binary channelized model containing more than 2 million voxels can be generated in about 12 seconds on a last generation intel\textsuperscript{\textregistered} i7 CPU, by randomly sampling a 125-dimensional uniform vector. Several 2D and 3D categorical TIs are used to study the unconditional simulation capabilities of our SGAN-based simulation approach. More importantly, synthetic inversion case studies involving 2D steady-state flow and 3D transient hydraulic tomography demonstrate the effectiveness of our SGAN for probabilistic inversion, with and without direct conditoning data. For the 2D case, the inversion rapidly explores the posterior model distribution, while for the 3D case the inversion produces model realizations that fit the data close to the target level and look similar to the true model. Main topics for future research include how to achieve direct conditioning to data for geostatistical simulation, improve simulation quality, accelerate training time, and adapt to continuous TIs.

\acknowledgments
Python codes of the proposed 2D/3D SGAN-based simulation and inversion approaches are available from the first author (and will be made available on \url{https://github.com/elaloy}). This work was partially supported by Agence Nationale de la Recherche through the grant ANR-16-CE23-0006 Deep in France. We thank the original 2D SGAN developers for sharing their code (\url{https://github.com/ubergmann/spatial_gan}). A temporary academic license of the DeeSse (DS) MPS code can be obtained upon request to one of its developers (Gr\'egoire Mariethoz, Philippe Renard, Julien Straubhaar). Finally, we thank Laurent Lemmens for sharing his PF and CF calculation routines.

{}

\end{document}